%% file: locality.tex
\documentclass[]{article}

\usepackage[english]{babel}
\usepackage{graphicx}
\usepackage{framed}
\usepackage[normalem]{ulem}
\usepackage{amsmath}
\usepackage{amsthm}
\usepackage{amssymb}
\usepackage{mathtools}
\usepackage{amsfonts}
\usepackage{ifthen}
\usepackage{hyperref}
\usepackage{widetext}
\usepackage[noadjust]{cite}
\newcommand{\citep}{\cite}
\newcommand{\TEVC}{need this command for conditional formatting to work.} 

\usepackage[utf8]{inputenc}
\usepackage[top=1 in,bottom=1in, left=1 in, right=1 in]{geometry}
\hypersetup{
    colorlinks=true,
    linkcolor=blue,
    filecolor=magenta,
    urlcolor=cyan,
}
\usepackage{enumerate, upgreek,mathtools}
\usepackage[lined,boxed,linesnumbered,noend]{algorithm2e}
\usepackage{tikz}

\newcommand{\shat}{\hat{s}}
\newcommand{\E}{\mathbb{E}}

\newcommand{\N}{\mathbb{N}}

\newcommand{\binstr}[1]{\{0,1\}^{#1}}

\theoremstyle{definition}
\newtheorem{theorem}{Theorem}
\newtheorem{corollary}[theorem]{Corollary}
\newtheorem{claim}[theorem]{Claim}
\newtheorem{definition}{Definition}
\newtheorem*{example}{Example}

\newtheorem{lemma}[theorem]{Lemma}

\newcommand{\defn}{\stackrel{\textrm{\clap{\scriptsize def}}}{=}}
\newcommand{\cond}[2]{\ifdefined\TEVC
                            #1\else
                            #2\fi }

\begin{document}
%
\title{Revisiting Locality in Binary-Integer Representations}
%
%
%

\author{Hrishee Shastri~and
        Eitan~Frachtenberg
\thanks{Department of Computer Science, Reed College, Oregon}}
\date{}

\maketitle

\begin{abstract}

Mutation and recombination operators play a key role in determining the speed and quality of Genetic and Evolutionary Algorithms (GEAs). 
Prior work has analyzed the effects of these operators on genotypic variation, often using locality metrics that measure the sensitivity and stability of genotype-phenotype representations to these operators.
In this paper, we focus on an important subset of representations, namely nonredundant bitstring-to-integer representations, and analyze them through the lens of Rothlauf's widely used locality metrics.
                                                                                         
We first define locality metrics equivalent to Rothlauf's that are tailored to our domain: the \textit{point locality} for single-bit mutation and \textit{general locality} for recombination.
With these definitions, we derive tight bounds and a closed form expected value for point locality. For general locality we show that it is asymptotically equivalent across all representations and operators.
We also recreate three established GEA experiments to understand the predictive power of point locality on GEA performance, focusing on two popular and often juxtaposed representations: standard binary and binary reflected Gray.

We show that standard binary has provably no worse locality than any Gray encoding, including binary reflected Gray. We discuss this result in the context of previous studies that found binary reflected Gray to outperform standard binary, and we argue that locality cannot be the explanation for strong performance. Finally, we provide empirical evidence that weak point locality representations can be beneficial to performance in the exploration phase of the GEA, while strong point locality representations are more beneficial in the exploitation phase. 

\end{abstract}


\section{Introduction}
\label{sec:intro}

Genetic and Evolutionary Algorithms (GEAs) solve optimization and search problems by codifying a population of possible solutions, evaluating their fitness in solving the problem, and iteratively modifying them in an attempt to improve their fitness. The digital manifestation of the solutions are called genotypes, and their interpretations into the specific problem domain are called phenotypes. The function that maps from genotypes to phenotypes is simply called the \emph{representation}, and it can have a significant impact on the success and speed of the GEA to approximate optimal solutions. Consequently, many empirical and theoretical studies investigated the effects of representation on GEA performance under different operators that modify genotypes. The No Free Lunch Theorems can be interpreted to show that no single representation has a performance advantage for all optimization problems~\citep{wolpert97:freelunch}.

A key component of evolution, both biological~\citep{mitchell07:evolutionary} and computational~\citep{goldberg92:genetic}, is the variation of genotypes. Perhaps the most common operator for variation in GEAs is mutation, which may be combined with another operator, recombination~\citep{back2018:evolutionary}. There are various implementations of the mutation operator, but typically they embody a localized change to an individual genotype. Contrast this with recombination, which requires two or more individuals and often involves nonlocal changes to the genotypes.

Perhaps the most commonly used variation operator is point mutation, a simple mutation operator which randomly changes one allele at a time. There exist various forms and parameters for point mutation that control the magnitude of the genotypic change. By controlling this magnitude, mutation can be used in a search GEA both for exploration---sampling many disparate parts of the search space---and for exploitation---thoroughly searching in a localized subspace~\citep{crepinsek13:exploration}. 

An important property of mutation that can increase the predictability and interpretability of the GEA is to have ``strong locality,'' which we informally define here as the property that small variations in the genotype lead to small variations in the phenotype~\citep{pereira06:analysis}. Strong locality implies better control of the GEA, because tuning mutation for a certain magnitude of changes in the genotype---the inputs to the search---leads to an expectation of the magnitude of changes in the phenotype---the outcome of the search. Note that the mutation operator is intricately tied to the representation. It is the combination of the mutation operator and representation that determines the magnitude of phenotypic change. Often, the discussion of locality assumes a fixed mutation operator---such as uniform bit flips or Gaussian differences---and focuses on the representations, such as standard binary or Gray.

In his seminal book,
\cond{Rothlauf \citep{rothlauf06:representations}}{\citet{rothlauf06:representations}} presented a theoretical framework for comparing representations based on three properties: redundancy, scaling, and locality. He proposed two metrics to quantify the locality of representations, one specifically for point mutation (simply called ``locality'' in the book) and one for any variation operator (called ``distance distortion'').

This paper focuses on these metrics when applied to a widely used subclass of representations, namely nonredundant translations from bitstring genotypes to nonnegative integer phenotypes, such as Gray encoding~\citep{whitley99:free}. These representations are very common in practice, and are also be used to represent fixed-point values and vectors of values.

\subsubsection*{Summary of contributions and paper outline}

After surveying related work in Section \ref{sec:related}, we examine the theoretical properties of both locality metrics. We build on Rothlauf's definitions for locality and distance distortion and redefine them specifically for the domain of binary-integer representations, which we call point locality and general locality, respectively. This lets us contribute proofs and precise computations for the following properties:

\begin{itemize}
    \item Tight lower and upper bounds for point locality.
    \item Showing that the point localities of both standard binary (SB) and binary-reflected Gray (BRG) encodings are identical and optimal.
    \item An existence proof and construction algorithm for suboptimal Gray encodings.
    \item A computation of the expected value of the point locality of random representations.
    \item A lower bound and asymptotic limit for general locality.
\end{itemize}

Sections \ref{sec:theory-point} and \ref{sec:theory-general} present these results in order, at higher detail than in previous work~\citep{shastri20:locality}.
To put these theoretical results in context, we also attempt in Sec.~\ref{sec:experiments} to faithfully reproduce three distinct past experiments from three different types of GEAs that empirically compared SB's performance to BRG's. These studies found BRG to generally outperform SB, and in some cases hypothesized that this better performance is the result of stronger locality~\citep{hinterding95:nature, rothlauf06:representations}. Here, we recreate these established findings of superior BRG performance and then examine them in the newly proven perspective of the equivalent locality of the two representations. Our experiments are therefore by design not new,  but rather a framework to delve deeper into the reasons for performance differences between representations.

Then, in Sec.~\ref{sec:discussion}, we discuss these results and produce additional experimental data and alternative explanations to BRG's superior performance. A separate contribution of this section is a discussion of the role of the GEA's simulation time on dynamically varying the representation, showing that different phases of the simulation can benefit from different locality properties of the representation. Finally, we conclude with suggestions for future work in Sec.~\ref{sec:conclusions}.

\section{Related work}
\label{sec:related}

The representation in a GEA is tightly linked to its performance, which led to numerous studies working to formalize and measure the effects of representation locality~\citep{galvan11:defining, gottlieb01:prufer, gottlieb99:characterizing, jones95:fitness, manderick91:genetic, ronald97:robust}. 
Many of these studies quantify locality in different ways and apply it to different phenotype classes, such as floating-point numbers~\citep{pereira06:analysis}, computer programs~\citep{galvan11:defining, rothlauf06:grammar}, permutations~\citep{galvan09:effects}, and trees~\citep{hoai06:representation, rothlauf99:tree}. 
The focus of these approaches is often to measure the effect of genotypic changes on fitness distances~\citep{gottlieb99:characterizing}. A more general approach is to instead measure the effect on phenotypic distances, because it provides a way to measure the locality of a representation that is independent of the fitness function~\citep{rothlauf03:locality}.

The foundational locality definitions for our study and several others~\citep{chiam06:issues, galvan10:towards} come from Rothlauf's treatise on the theory of GEA representations \citep{rothlauf06:representations}. Rothlauf defines the locality $d_m$ as:

\begin{equation} \label{eq:rothlauf-locality}
d_m \defn \sum_{d_{x,y}^{g}=d_{min}^{g}}|d_{x,y}^{p}-d_{min}^{p}|~~~\text{(ibid. Eq. 3.23)},
\end{equation}

where for every two distinct genotypes $x,y$, $d_{x,y}^{g}$ is the genotypic distance between $x$ and $y$, and $d_{x,y}^{p}$ is the phenotypic distance between their respective phenotypes, based on our choices of genotypic and phenotypic spaces. Similarly, $d_{min}^{g}$ and $d_{min}^{p}$ represent the minimum possible distance between genotypes or phenotypes, respectively.\footnote{This definition has changed from the book's first edition to better match reader intuition and for simpler computation.}
For example, for nonredundant representations of integers as bitstrings (the focus of this paper), genotypic distances are measured in Hamming distance and phenotypic distances use the usual Euclidean metric in $\mathbb{N}$.

This definition ``describes how well neighboring genotypes correspond to neighboring phenotypes'' (ibid. p. 77), which is valuable for measuring small genotypic changes that typically result from mutation. Extending this notion to include large genotypic changes (e.g., from a recombination operator), Rothlauf defines the \emph{distance distortion} $d_c$ as:

\begin{equation} \label{eq:rothlauf-distancedistortion}
d_c \defn \frac{2}{n_{p}(n_{p}-1)}\sum_{i=1}^{n_{p}}\sum_{j=i+1}^{n_{p}}|d^{p}_{x_{i},x_{j}}-d^{g}_{x_{i},x_{j}}|~~~\text{(ibid. Eq. 3.24)},
\end{equation}

where $n_{p}$ is the size of the search space, and $d^{p}_{x_{i},x_{j}}, d^{g}_{x_{i},x_{j}}$ are the phenotypic and genotypic distance, respectively, between two distinct individuals $x_{i}$, $x_{j}$. The term $\frac{2}{n_{p}(n_{p}-1)}$ is equal to $\frac{1}{\binom{n_{p}}{2}}$, the proportion of each distinct pair of individuals. This definition ``describes how well the phenotypic distance structure is preserved when mapping $\Phi_{p}$ on $\Phi_{g}$'', where $\Phi_{p}$ is the phenotypic search space and $\Phi_{g}$ is the genotypic search space (ibid.~p. 84).

As observed by \cond{Galván-López \citep{galvan11:defining}}{\citet{galvan11:defining}}, high values of $d_m$ and $d_c$ actually denote low locality while low values denote high locality. To avoid confusion, we will refer to low metric values as strong locality and high metric values as weak locality.

Similar in spirit, \cond{Gottlieb and Raidl \citep{gottlieb99:characterizing}}{\citet{gottlieb99:characterizing}} also defined a pair of locality metrics for mutation and crossover operators called \emph{mutation innovation} and \textit{crossover innovation}.
They additionally defined \emph{crossover loss}, which measures the number of phenotypic properties that are lost by crossover. These metrics are probabilistic and empirical in nature, so they are harder to reason about analytically. But they have been demonstrated in practice to predict GEA performance on the multidimensional knapsack problem~\citep{pereira06:analysis, raidl05:empirical}. 

In a different study, \cond{Chiam, Goh, and Tan \citep{chiam06:issues}}{\citet{chiam06:issues}} defined the concept of \emph{preservation}, which ``measure[s] the similarities between the genotype and phenotype search space.'' Their study uses Hamming distance between genotypes and L2 norms between phenotypes to define analogous metrics to Rothlauf's (called \emph{proximity preservation} and \emph{remoteness preservation}). Unlike Rothlauf's metrics, their metrics look in both directions of the genotype-phenotype mapping. The authors demonstrated with examples (as we prove formally in the next section) that SB and BRG have the same genotype-to-phenotype locality, but not phenotype-to-genotype locality. They also predicted, based on this similarity, that crossover-based GEAs would perform about the same with both SB and BRG encodings, which appears to contradict some past experimental results, as we discuss in Sec.~\ref{subsec:GA}.

A different approach to approximating locality was given by \cond{Mathias and Whitley \citep{mathias94:transforming}}{\citet{mathias94:transforming}}, and independently, by \cond{Weicker \citep{weicker10:binary}}{\citet{weicker10:binary}} using the various norms of the phenotype-genotype distance matrix for different representations. These studies did not draw performance predictions from these metrics.

Another interesting aspect of locality of representation is its effect over time.
A number of studies explored the possibility of changing the representation dynamically during the GEA execution, primarily to avoid premature convergence. 
For any function, there are multiple representations that make the optimization problem trivial~\citep{liepins90:representational}, so it could make sense to try different representations dynamically.
Whitley, Rana, and Heckendorn even computed the expected number of local optima for a random representation~\citep{whitley97:representation}.
The first two authors also introduced the concept of \emph{shifting} to dynamically switch from one Gray representation to another to escape local optima~\citep{rana97:bit}.
\cond{Barbulescu, Watson, and Whitley \citep{barbulescu00:dynamic}}{\citet{barbulescu00:dynamic}} expanded this work by proving a bound on the number of available Gray representations available via shifting. They showed that to improve performance, the GEA needs to shift to a dissimilar representation to the current one.

Shifting between representations occurs in sequence in these works, but can also be evaluated in parallel using a modified island model~\citep{skolicki04:improving}.
In all these cases, however, the different representations are either standard binary or Gray. These studies have not explored the possibility of changing to a random representation with different locality to evaluate its effect on GEA performance.

\cond{Whitley \citep{whitley00:functions}}{\citet{whitley00:functions}} did compute summary statistics for any random permutation function, but these did not include average locality.
To offer a framework of understanding the effect of arbitrary representations on locality, we also compute in Sec.~\ref{subsec:expected_locality} the expected point locality of any random representation.


\section{Theoretical results on point locality}
\label{sec:theory-point}

\subsection{Defining point locality}
\label{subsec:point-defintion}

Recall Rothlauf's definition for locality (Eq.~\ref{eq:rothlauf-locality}). Since we are for the moment specifically concerned with single-bit mutations, both $d^{g}_{min}$ and $d^{p}_{min}$ are equal to 1. And since we limit ourselves to binary-integer representations, all phenotypic distances fall in the range $[0,2^{\ell}-1]$ and all genotypic distances fall in the range $[0,\ell]$. We can use these assumptions to precisely define our own derived metric, \emph{point locality}, using units that we find more intuitive.

We first define a representation $r : \{0,1\}^{\ell} \rightarrow [0,2^{\ell})$ as a bijection between the set of $\ell$-bit bitstrings $\{0,1\}^{\ell}$ and the discrete integer interval $[0,2^{\ell})$. This ensures that the representation is not redundant -- i.e., every integer in the interval $[0,2^{\ell})$ is represented by exactly one $\ell$-bit bitstring, and the number of search-space points $n_p$ is exactly $2^{\ell}$. A representation $r$ can therefore be equivalently described as a permutation $\pi: [0,2^{\ell}) \rightarrow [0,2^{\ell})$, where $\pi (i) = j$ if and only if the SB representation of $i$ maps to $j$ under $r$. Consequently, we can write $\pi$ as a $2^{\ell}$-tuple where the $i^{th}$ coordinate (starting at 0) is $\pi (i)$. We also use the notation $\shat_{i}$ to denote the binary string produced from flipping the $i^{\text{th}}$ coordinate of the binary string $s \in \{0,1\}^{\ell}$. Formally, $\shat_{i}=s\oplus{}2^{i}$, where $\oplus$ denotes bitwise exclusive-or.

We can now define the \emph{point locality} $p_r$ for a nonredundant bitstring-to-integer representation $r$ as the average change in phenotypic value for a uniformly random single-bit flip in the genotype. More formally:

\begin{definition} 
The point locality $p_{r}$ for a representation $r$ is $p_r \defn \overline{|r(\shat_{i}) - r(s)|} \quad  \forall s \in \{0,1\}^{\ell}, \quad \forall i \in [0,\ell)$. Explicitly, 
    \begin{equation} \label{point-locality}
        p_{r} \defn  \frac{\sum_{s}\sum_{i}|r(\shat_{i}) - r(s)|}{2^{\ell}\cdot \ell}.
    \end{equation}
\end{definition}

Note that for any given $\ell$, our definition of $p_r$ is a simple linear transformation of Rothlauf's $d_m$. In our domain, $d_m$ simply sums the phenotypic distances minus one between all distinct genotypic neighbors, while $p_{r}$ computes the average phenotypic distance between all ordered pairs of genotypic neighbors. Also note that in Rothlauf's $d_{m}$, $d^{p}_{min}$ occurs in the summation $\ell 2^\ell$ times. Coupled with the fact that $d^{p}_{min} = 1$ in our domain, we have the relationship:
\begin{align*}
                p_{r} &= \frac{2d_{m}+ \ell 2^{\ell}}{{\ell}2^{\ell}} 
                  = \frac{d_{m}}{{\ell}2^{\ell-1}} + 1.
\end{align*}


\subsection{Computing tight bounds for point locality}
\label{subsec:point-bounds}

Our first analysis proves lower and upper bounds on $p_{r}$, computes $p_r$ for SB and Gray representations, and verifies the existence of representations with both minimum and maximum $p_{r}$. 

\subsubsection{Lower bound}

\begin{theorem}[Lower bound] \label{lr-lowerbound}
        $p_r \geq \frac{2^{\ell} - 1}{\ell}$.
\end{theorem}

\begin{proof}
    We reduce the problem of minimizing locality to another problem, that of enumerating nodes on a hypercube while minimizing neighbor distances, for which lower and upper bounds have been established by \cond{Harper \citep{harper64:optimal}}{\citet{harper64:optimal}}.
    We use the term $\Delta_{ij} \defn |i - j|$ to denote the absolute difference between the numbers assigned to two adjacent vertices $i$ and $j$ on the unit $\ell$-cube. The unit $\ell$-cube consists of all elements in $\{0,1\}^{\ell}$. Two vertices in the $\ell$-cube are adjacent if they differ by only one bit (i.e., they have Hamming distance 1). Note that assigning a number $n$ to a vertex $i$ can be thought of as a representation $r$ mapping $i$ to $n$, or $r(i) = n$. Therefore, we have $\Delta_{\shat_{i}s} = |r(\shat_{i}) - r(s)|$
    for adjacent vertices $\shat_{i}$ and $s$, since a 1-bit difference is equivalent to a single bit-flip mutation.
    Following \cond{Harper}{\citet{harper64:optimal}}, we define the sum $\sum \Delta_{\shat_{i}s}$  to be the sum of the absolute difference between two adjacent vertices $\shat_{i}$ and $s$ that runs over all possible pairs of neighboring vertices in the $\ell$-cube. Note that
    \begin{align*}
        2\sum \Delta_{\shat_{i}s} &= \sum_{s}\sum_{i}|r(\shat_{i}) - r(s)|, 
    \end{align*}
    since the RHS computes $|r(\shat_{i}) - r(s)|$ twice for every ordered pair.
    
    Harper proved that $\sum \Delta_{\shat_{i}s} \geq 2^{\ell-1}(2^{\ell}-1)$. Therefore, 
    \begin{align*}
        p_r &= \frac{\sum_{s}\sum_{i}|r(\shat_{i}) - r(s)|}{2^{\ell}\cdot \ell} \\
            &= \frac{2\sum \Delta_{\shat_{i}s}}{2^{\ell}\cdot \ell} \\
            &\geq \frac{2(2^{\ell-1}(2^{\ell}-1))}{2^{\ell}\cdot \ell} \quad (ibid.) \\
            &= \frac{2^{\ell}-1}{\ell},
    \end{align*}
    proving that $p_r \geq \frac{2^{\ell}-1}{\ell}$.
\end{proof}

\begin{corollary} \label{SB-opt}
    Standard binary encoding is optimal, meaning that it has the strongest point locality equal to $\frac{2^{\ell} - 1}{\ell}$.
\end{corollary}
\begin{proof}
   SB encoding is also a representation---call it $SB$. We consider $p_{SB}$: 
    \begin{align*}
        p_{SB} &= \frac{\sum_{s}\sum_{i}|SB(\shat_{i}) - SB(s)|}{2^{\ell}\cdot \ell}.
    \end{align*}
    The inner sum $\sum_{i}|SB(\shat_{i}) - SB(s)|$ computes the sum of all differences obtained from flipping the $i^{th}$ bit of a given SB string $s$. Flipping the $i^{th}$ bit elicits an absolute phenotypic difference of $2^i$ for any $s$ and $i$, reducing the inner sum to:
    \begin{align*}
        \sum_{i}|SB(\shat_{i}) - SB(s)| &= \sum_{i=0}^{\ell-1}2^i = 2^{\ell} - 1. 
    \end{align*}
    Now since there are $2^\ell$ elements in $\{0,1\}^\ell$, the outer sum reduces to $\sum_{s}(2^{\ell}-1) = (2^{\ell})(2^{\ell}-1)$. Combining these lets us compute $p_{SB}$:
    \begin{align*}
         p_{SB} &= \frac{\sum_{s}\sum_{i}|SB(\shat_{i}) - SB(s)|}{2^{\ell}\cdot \ell} \\
                &= \frac{2^{\ell}(2^{\ell}-1)}{2^{\ell}\cdot \ell} \\
                &= \frac{2^{\ell}-1}{\ell},
    \end{align*}
    which is the lower bound given by Theorem \ref{lr-lowerbound}. Thus, SB has optimal point locality.
\end{proof}

\subsubsection{Upper bound}

\begin{theorem}[Upper bound] \label{lr-upperbound}
    $p_r \leq 2^{\ell-1}$.
\end{theorem}
\begin{proof}
    We rely on another result from
    \cond{Harper \citep{harper64:optimal}}{\citet{harper64:optimal}} in which he proved
    that $\sum \Delta_{\shat_{i}s} \leq {\ell}2^{2(\ell-1)}$. We have 
    \begin{align*}
            p_r &= \frac{\sum_{s}\sum_{i}|r(\shat_{i}) - r(s)|}{2^{\ell}\cdot \ell} \\
                &= \frac{2\sum \Delta_{\shat_{i}s}}{2^{\ell}\cdot \ell} \\
                &\leq \frac{2({\ell}2^{2(\ell-1)})}{2^{\ell}\cdot \ell} \quad (ibid.) \\
                &= \frac{2^{2(\ell-1)}}{2^{\ell-1}}  \\
                &= 2^{\ell-1}.
    \end{align*}
    Thus $p_{r} \leq 2^{\ell-1}$.
\end{proof}

\begin{claim} \label{worst-rep}
    There exists a representation $r$ with upper bound point locality $p_{r} = 2^{\ell-1}$.
\end{claim}
\begin{proof}
     Here, we reduce the problem of constructing a representation $r$ with upper bound locality $p_{r} = 2^{\ell-1}$ to that of assigning integers in $[0,2^{\ell})$ to vertices in the $\ell$-cube, such that $\sum \Delta_{\shat_{i} s}$ is maximized. \cond{Harper \citep{harper64:optimal}}{\citet{harper64:optimal}} constructed an algorithm assigning numbers to vertices to maximize $\sum \Delta_{\shat_{i} s}$, shown in Algorithm \ref{alg:harper-upperbound}.
   Maximizing $\sum \Delta_{\shat_{i} s}$ is equivalent to maximizing $p_{r}$, so such a representation exists.

\begin{algorithm}
  $V \gets \{0,1\}^{\ell}$\;
  $s \gets \text{randomly selected node from $V$}$\;
  $\text{assign($s$,0)}$\;
  $\text{parity} \gets |s|_{1} \text{mod } 2$\;
  $L_{1} \gets \text{randomly shuffled } [1,...,2^{\ell - 1} -1]$\;
  $L_{2} \gets \text{randomly shuffled } [2^{\ell - 1},...,2^{\ell}-1]$\;
  $i \gets 0$\;
  $j \gets 0$\;
  \For{$v \in V\setminus\{s\}$}
  {
    \uIf{$|v|_{1} \text{mod } 2 = \text{parity}$}{
        $\text{assign($v$,$L_{1}[i]$)}$ \;
        $i \gets i + 1$\;
      }
      \Else{
        $\text{assign($v$,$L_{2}[j]$)}$ \;
        $j \gets j + 1$\;
      }
  }
  \caption{Harper's algorithm to assign integers on the $\ell$-cube to maximize $\sum \Delta_{\hat{s}_{i},s}$, the sum of neighbor differences. The function assign($v,n$) assigns the number $n$ to $\ell$-cube vertex $v$. $|v|_{1}$ denotes the number of `1' bits in vertex $v$.}\label{alg:harper-upperbound}
\end{algorithm}

\end{proof}


\subsection{Computing point locality for Gray encodings}
\label{subsec:gray-point}

Gray encodings are encodings where two phenotypic neighbors are also genotypic neighbors. In our domain, this means that any two phenotypes that are separated by one unit are represented by two genotypes that are separated by one bit flip. For a given length $\ell$, there are multiple Gray encodings, but the most common is BRG. Here, we show that BRG has equivalent point locality to SB, and that not all Gray encodings share the same locality value.

\subsubsection{Point locality of BRG}

\begin{claim} \label{BRG-opt}
    Binary Reflected Gray (BRG) encoding is also optimal.
\end{claim}
\begin{proof}
   Let $BRG$ notate the representation for Binary Reflected Gray and \\
   \begin{align*}
        p_{BRG} &= \frac{\sum_{s}\sum_{i}|BRG(\shat_{i}) - BRG(s)|}{2^{\ell}\cdot \ell}.
   \end{align*}
   
   We start by proving the following two lemmas: 
   
   \begin{lemma} \label{lemma1}
   $\sum_{s \in \binstr{\ell}}|BRG(\shat_{\ell-1}) - BRG(s)| = 2^{2\ell-1}$.
   \end{lemma}
   In other words, the sum of the differences obtained by flipping the leftmost bit over all $\ell$-bit bitstrings in BRG encoding is $2^{2\ell-1}$.   
   \begin{proof}
      Consider the recursive nature of BRG codes \citep{rowe04:properties}. Let $L_{\ell}$ be the ordered list of $\ell$-bit BRG codes where $L_{\ell}[i]$ is the bitstring that maps to $i$. Note that $2^{\ell}$ is the length of $L_{\ell}$ and $[\hspace{1mm}]$ denotes list indexing. The left half of $L_{\ell}$ contains $L_{\ell-1}$ prefixed with 0 and the right half of $L_{\ell}$ contains $L_{\ell-1}$ in reverse order prefixed with 1. Flipping the leftmost bit of $L_{\ell}[i]$ will yield $L_{\ell}[2^{\ell}-1-i]$. Thus we have 
      \begin{align*}
          \cond{&}{}\sum_{s \in \binstr{\ell}}{|BRG(\shat_{\ell-1}) - BRG(s)|} \cond{\\}{}
                &=  \sum_{i=0}^{2^{\ell}-1}{|BRG(L_{\ell}[2^{\ell}-1-i]) - BRG(L_{\ell}[i])|} \\
                &= \sum_{i=0}^{2^{\ell}-1}{|2^{\ell}-1-i-i|} \\
                &= \sum_{i=0}^{2^{\ell}-1}{|2^{\ell}-(2i+1)|}. 
      \end{align*}
      Note that for a given $i \in [0,2^{\ell}-1]$, \\
                                                \[|2^{\ell}-(2i+1)| =\begin{cases} 
                                                      2^{\ell}-(2i+1) &  i < 2^{\ell}/2  \\
                                                      (2i+1)-2^{\ell} &  i \geq 2^{\ell}/2,  \\
                                                   \end{cases}
                                                \] \\
    which lets us split the sum to 
    \begin{align*}
        &= \sum_{i=0}^{2^{\ell-1}-1}\left(2^{\ell}-(2i+1)\right) + \sum_{i=2^{\ell-1}}^{2^{\ell}-1}\left({(2i+1)-2^{\ell}}\right) \\
        &= \sum_{i=0}^{2^{\ell-1}-1}{2^{\ell}} - \sum_{i=0}^{2^{\ell-1}-1}{(2i+1)}
        + \sum_{\mathclap{i=2^{\ell-1}}}^{2^{\ell}-1}{(2i+1)} - \sum_{\mathclap{i=2^{\ell-1}}}^{2^{\ell}-1}{2^{\ell}} \\
        &= (2^{\ell-1})(2^{\ell}) - \sum_{i=0}^{\mathclap{2^{\ell-1}-1}}{(2i+1)}
        + \sum_{\mathclap{i=2^{\ell-1}}}^{2^{\ell}-1}{(2i+1)} - (2^{\ell-1})(2^{\ell}) \\
        &= \sum_{i=2^{\ell-1}}^{2^{\ell}-1}{(2i+1)}- \sum_{i=0}^{2^{\ell-1}-1}{(2i+1)}.    
    \end{align*}
    Splitting the left sum and using the facts that the sum of the first $n$ odd numbers is $n^{2}$,
    \begin{align*}
        &= \left(\sum_{i=0}^{2^{\ell}-1}(2i+1) - \sum_{i=0}^{2^{\ell-1}-1}(2i+1)\right) - (2^{\ell-1})^{2} \\
        &= (2^{\ell})^{2} - (2^{\ell-1})^{2} - (2^{\ell-1})^{2} \\
        &= 2^{2\ell} - 2\cdot 2^{2\ell-2} \\
        &= 2^{2\ell} - 2^{2\ell-1} \\
        &= 2^{2\ell}(1-\frac{1}{2}) \\
        &= 2^{2\ell-1},
    \end{align*}
    Thus $\sum_{s \in \binstr{\ell}}|BRG(\shat_{\ell-1}) - BRG(s)| = 2^{2\ell-1}$. 
   \end{proof}
   
   \begin{lemma} \label{lemma2}
   $\sum_{s}\sum_{i}|BRG(\shat_{i}) - BRG(s)| = 2^{2\ell} - 2^\ell$.
   \end{lemma}
    \begin{proof}
    We proceed with induction on $\ell$. For the base case ($\ell=1$), the set $\binstr{1}$ contains two BRG codes, $\{0,1\}$, which corresponds to the integers 0 and 1, respectively. Thus $\sum_{s \in \binstr{1}}\sum_{i=0}^{1-1}|BRG(\shat_{i}) - BRG(s)| = 1 + 1 = 2 = 2^{2\cdot 1} - 2^1$.
    
    For the inductive hypothesis (I.H.), assume $\sum_{s}\sum_{i}|BRG(\shat_{i}) - BRG(s)| = 2^{2\ell} - 2^\ell$ for some $\ell \in \N$.
            We must now show that $\sum_{s \in \binstr{\ell+1}}\sum_{i=0}^{\ell}|BRG(\shat_{i}) - BRG(s)| = 2^{2(\ell+1)} - 2^{(\ell+1)}$. Note that in the inductive step, we are working with strings of length $\ell+1$.
            \begin{align*}
                &\sum_{s \in \binstr{\ell+1}}\sum_{i=0}^{\ell}|BRG(\shat_{i}) - BRG(s)|& \\ 
                &= \sum_{s \in \binstr{\ell+1}}\left(|BRG(\shat_{\ell}) - BRG(s)| + \sum_{i=0}^{\ell-1}|BRG(\shat_{i}) - BRG(s)|\right) \\
                &= \sum_{s \in \binstr{\ell+1}}\sum_{i=0}^{\ell-1}|BRG(\shat_{i}) - BRG(s)| + \sum_{\mathclap{s \in \binstr{\ell+1}}}|BRG(\shat_{\ell}) - BRG(s)|. 
            \end{align*}
            By I.H. and the fact that there are two copies of the $\ell$-bit BRG code in the $(\ell+1)$-bit BRG code,
            \begin{align*}
                &= 2\cdot (2^{2\ell} - 2^{\ell}) + \sum_{s \in \binstr{\ell+1}}|BRG(\shat_{\ell}) - BRG(s)|. 
            \end{align*}
            By Lemma~\ref{lemma1},
            \begin{align*}
                    &= 2\cdot (2^{2\ell} - 2^{\ell}) + 2^{2(\ell+1)-1} \\
                                        &= 2^{2\ell+1} - 2^{\ell+1} + 2^{2\ell+1} \\
                                        &= 2\cdot 2^{2\ell+1} - 2^{\ell+1} \\
                                        &= 2^{2(\ell+1)} - 2^{(\ell+1)}.
            \end{align*}
    \end{proof}

    Now we can prove Corollary~\ref{BRG-opt}. Considering $p_{BRG}:$ 
    \begin{align*}
        p_{BRG} &= \frac{\sum_{s}\sum_{i}|BRG(\shat_{i}) - BRG(s)|}{2^{\ell}\cdot \ell} \\
                &= \frac{2^{2\ell}-2^{\ell}}{2^{\ell} \cdot \ell} \quad \text{Lemma~\ref{lemma2}} \\
                &= \frac{2^{\ell}(2^{\ell}-1)}{2^{\ell} \cdot \ell} \\
                &= \frac{2^{\ell} - 1}{\ell},
    \end{align*}
    which is the lower bound given by Theorem \ref{lr-lowerbound}. Therefore, BRG has optimal point locality.
\end{proof}

Note that this equivalence in point locality between SB and BRG has already been demonstrated empirically for small values of $\ell$ \citep{chiam06:issues}, but our proof holds for all values of $\ell$.

\subsubsection{Suboptimal Gray encodings}

We now shift to generating a Gray encoding with suboptimal point locality, which requires the following:

\begin{lemma} \label{hypercube-hamiltonian}
For $\ell \geq 3$, the $\ell$-cube $Q_{\ell}$ always contains a Hamiltonian path starting with the sequence of nodes $0^{\ell -3}000$, $0^{\ell - 3}001$, $0^{\ell - 3}011$, $0^{\ell - 3}111$, where the notation $0^{x}$ denotes a length-$x$ bitstring of all 0s.
\end{lemma}
\begin{proof}
    Recall that the $\ell$-cube contains $2^{\ell}$ vertices labeled as binary strings in $\{0,1\}^{\ell}$, where vertex $i$ is connected to vertex $j$ if and only if $i$ and $j$ have Hamming distance 1. We proceed by induction on $\ell$.
    
    For the base case ($\ell = 3$), the path $000, 001, 011, 111, 101, 100, 110, 010$ is Hamiltonian. 
    For the inductive step, assume that $Q_{\ell}$ has a Hamiltonian path starting with  $0^{\ell -3}000$, $0^{\ell - 3}001$, $0^{\ell - 3}011$, $0^{\ell - 3}111$. 
    Consider $Q_{\ell + 1}$. 
    By the inductive step, we can trace a path starting with the sequence $0^{\ell + 1}$, $0^{\ell-2}001$, $0^{\ell -2}011, 0^{\ell - 2}111$ to some node $0v$ ($v \in \{0,1\}^{\ell})$ such that the first bit never flips to $1$. From there, we hop to $1v$, then trace out the remainder of the path until all nodes have been visited, as nodes of the form $1x$ for all $x \in \{0,1\}^{\ell}$ constitute another copy of $Q_{\ell}$.
    The resulting path is Hamiltonian, since every node has been visited exactly once. 
\end{proof}

\begin{claim} \label{nongreedygray}
    There exists a Gray encoding $g$ with suboptimal point locality $p_g > \frac{2^{\ell} -1}{\ell}$ for any $\ell \geq 3$.
\end{claim}
\begin{proof}
     A bitstring $b \in \{0,1\}^\ell$ has $\ell$ neighbors that are all Hamming distance one away. We can thus reduce the problem of constructing a Gray code to that of constructing a Hamiltonian path on the $\ell$-cube. Recall that in Theorem~\ref{lr-lowerbound} we mapped the problem of minimizing locality to that of minimizing $\sum \Delta_{\shat_{i}s}$. In the same paper,
    \cond{Harper \citep{harper64:optimal}}{\citet{harper64:optimal}}
    formulates an algorithm that provably generates all representations that minimize $\sum \Delta_{\shat_{i}s}$, which we describe in Algorithm \ref{alg:harper-lowerbound}:

\begin{algorithm}
  $V \gets \{0,1\}^{\ell}$\;
  $s \gets \text{randomly selected node from $V$}$\;
  $\text{assign($s$,0)}$\;
  $i \gets 1$\;
  \While{$i \leq 2^{\ell} - 1$}
  {
    $v \gets \text{unassigned node in $V$ with highest number of already assigned neighbors}$\; 
    \tcc{If there are multiple nodes with this property, choose one at random}
    $\text{assign($v$, $i$)}$\;
    $i \gets i + 1$\;
  }
  \caption{Harper's algorithm to assign integers on the hypercube $\{0,1\}^{\ell}$ to minimize $\sum \Delta_{\hat{s}_{i},s}$, the sum of neighbor differences. The function assign($v,n$) assigns the number $n$ to hypercube vertex $v$.}\label{alg:harper-lowerbound}
\end{algorithm}

    Our goal is to construct a Hamiltonian path that violates this algorithm. This path in turn will determine a Gray code that has suboptimal point locality, because Harper's algorithm generates all representations with optimal point locality.
    Our modified algorithm starts by assigning 0 to vertex $0^{\ell-3}000$. We then assign 1 to $0^{\ell-3}001$ and assign 2 to $0^{\ell-3}011$. Algorithm \ref{alg:harper-lowerbound} would force us to assign 3 to $0^{\ell-3}010$ if we wanted to produce an optimal Gray code. Instead, we assign 3 to $0^{\ell-3}111$, which violates the algorithm. The remainder of the path can be traversed arbitrarily such that it is Hamiltonian, due to Lemma \ref{hypercube-hamiltonian}. This construction generates a Gray code $g$ with $p_{g} > \frac{2^{\ell}-1}{\ell}$ for $\ell \geq 3$. 
\end{proof}

\begin{figure}
    \resizebox{.99\linewidth}{!}{\input{bad_gray}}
    \caption{%
        A visualization of our modified algorithm from claim \ref{nongreedygray} used to generate $g$. In the fourth step, we assign $3$ to $111$ instead of $010$, violating Algorithm \ref{alg:harper-lowerbound} and thus ensuring $g$ is a suboptimal Gray code.
    }
    \label{fig:bad-gray}
\end{figure}
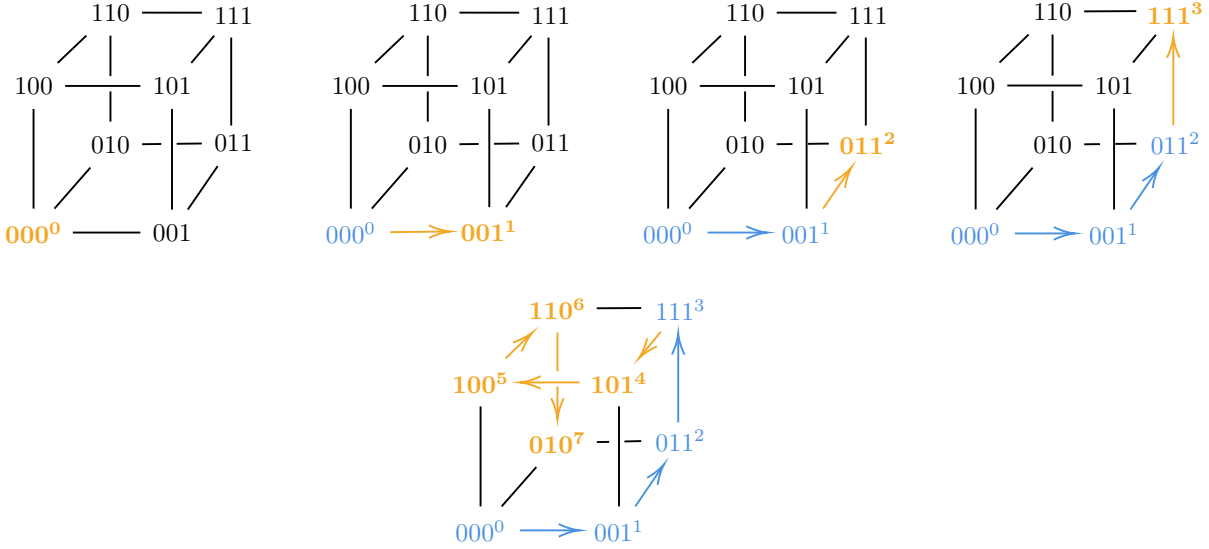

\begin{example}
    Consider a 3-bit Gray code representation $g$, given by the permutation $\pi = [0,1,3,7,5,4,6,2]$:
        \begin{align*}
            000 &\mapsto 0 \\
            001 &\mapsto 1 \\
            011 &\mapsto 2 \\
            111 &\mapsto 3 \\
            101 &\mapsto 4 \\
            100 &\mapsto 5 \\
            110 &\mapsto 6 \\
            010 &\mapsto 7.           
        \end{align*}
    Then $g$ is a Gray code generated by claim \ref{nongreedygray}, and $p_{g} \approx 3.667$ which is greater than the lower bound $\frac{2^{3}-1}{3} \approx 2.333$ from Theorem \ref{lr-lowerbound}. A visualization of how our modified algorithm from claim \ref{nongreedygray} generates $g$ appears in figure \ref{fig:bad-gray}. Note that in the fourth step, we assign $3$ to $111$, which has fewer neighbors than $010$ that are already labeled.
\end{example}

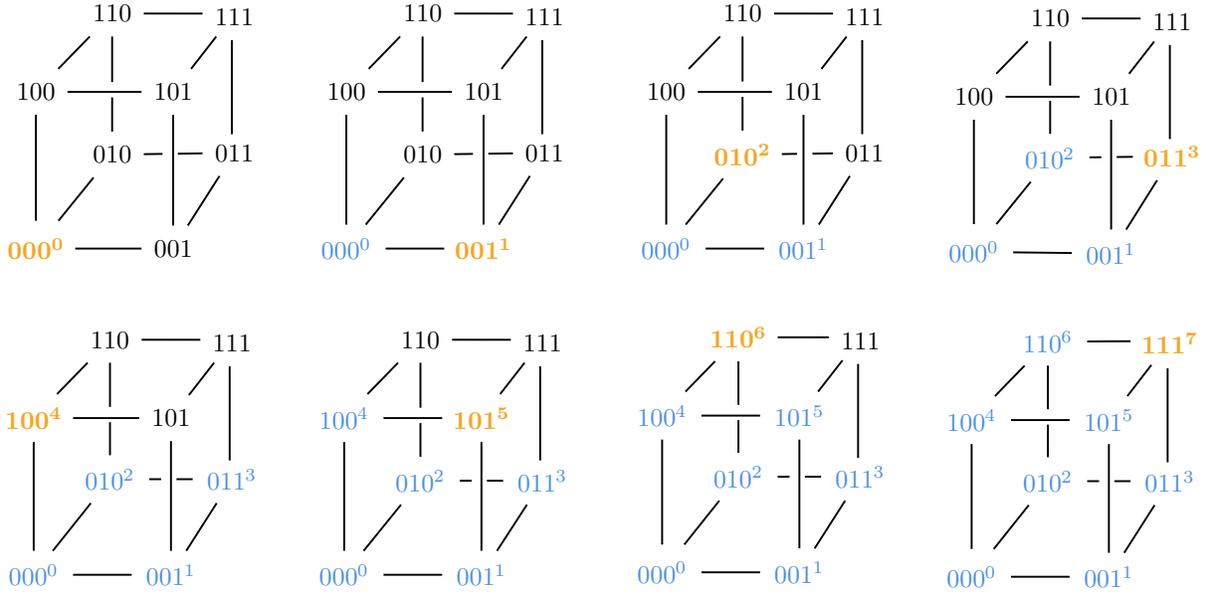
\begin{figure}
    \resizebox{.99\linewidth}{!}{\input{binary_harper}}
    \caption{%
        Algorithm \ref{alg:harper-lowerbound} generating SB for $\ell=3$.
    }
    \label{fig:binary_harper}
\end{figure}

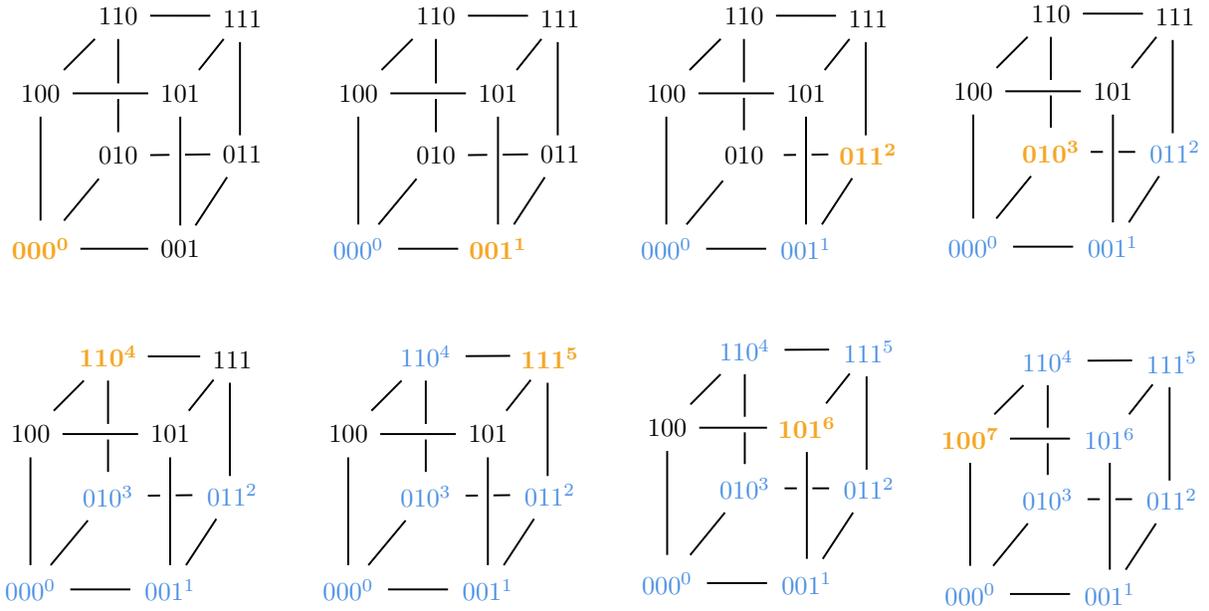
\begin{figure}
    \resizebox{.99\linewidth}{!}{\input{brg_harper}}
    \caption{%
        Algorithm \ref{alg:harper-lowerbound} generating BRG for $\ell=3$.
    }
    \label{fig:brg_harper}
\end{figure}

Also note that Algorithm \ref{alg:harper-lowerbound} can generate both SB and BRG codes, as illustrated in Figures \ref{fig:binary_harper} and \ref{fig:brg_harper}.

\subsection{Expected point locality}
\label{subsec:expected_locality}

Our final proof on point locality answers the question: ``what is the expected point locality of an arbitrary representation?''

\begin{theorem}[Expected Point Locality] \label{lr-average}
    $\E[p_{r}] = \frac{2^{\ell} + 1}{3}$.
\end{theorem}
\begin{proof}
    We begin with the definition of $p_r$: 
    \begin{align*}
        p_{r} &\defn \frac{\sum_{s}\sum_{i}|r(\shat_{i}) - r(s)|}{\ell 2^{\ell}} 
    \end{align*}
    \begin{equation}\label{start-pr-avg}
        \implies \E[p_{r}] = \frac{1}{\ell 2^{\ell}}\E\left[\sum_{s}\sum_{i}|r(\shat_{i}) - r(s)|\right].
    \end{equation}
    Let $L = \{0,...,2^{\ell} - 1\}$.
    Computing the expectation in equation \eqref{start-pr-avg} is equivalent to assigning numbers in $L$ to vertices of the $\ell$-cube uniformly at random, and then summing the absolute differences between every adjacent pair. 
    Let $k$ denote the value assigned to vertex $s$, and let $\pi_{k}$ denote a size $\ell$ random subset of $L \setminus \{k\}$. 
    
    Due to linearity of expectation, the sum of absolute differences for each vertex $s$ can be examined independently even though they are dependent variables. Then, $s$ has $\ell$ neighbors where the $i^{th}$ neighbor is assigned the $i^{th}$ element in $\pi_{ k}$, which we denote $\pi_{ k}(i)$.
    This gives us the following relation (letting $n = 2^{\ell}$ for simplicity):
    
    \begin{equation}\label{pr-avg-next}   \E\left[\sum_{s}\sum_{i}|r(\shat_{i}) - r(s)|\right] = \E\left[\sum_{k=0}^{n-1}\sum_{i = 1}^{\ell} | \pi_{k}(i) - k|\right].
    \end{equation}
    Since every number in $L \setminus \{k\}$ has equal probability of appearing as the $i$th element of $\pi_{ k}$, $\E[\pi_{k}(i)]$ does not depend on $i$ at all. Therefore, once we apply linearity of expectation on the right side of equation \eqref{pr-avg-next}, we can replace $\pi_{k}(i)$ with a discrete uniform random variable sampled from $L \setminus \{k\}$. 
    Denote this random variable as $X^{k}$.
    We then have:    
    \begin{align*}
        &= \ell\sum_{k = 0}^{n-1}\E[|X^{k}-k|] \\
        &= \ell\sum_{k = 0}^{n-1}\sum_{x \in L \setminus \{k\}} |x-k|\cdot \text{P}(X^{k} = x) \\
        &= \frac{\ell}{n-1}\sum_{k = 0}^{n-1}\sum_{x \in L \setminus \{k\}} |x-k| \\
        &= \frac{\ell}{n-1}\sum_{k = 0}^{n-1}\left( S_{1,k} + S_{1, n-k-1}\right), 
    \end{align*}
    where we let $S_{i,j} = \sum_{m = i}^{j}m$. Continuing with the fact that $S_{1,m}  = \frac{m(m+1)}{2}$:
    \begin{align*}
        &= \frac{\ell}{2(n-1)}\sum_{k = 0}^{n-1}\left((n-k-1)(n-k)+k(k+1)\right) \\
        &= \frac{\ell}{2(n-1)}\left(\sum_{k=0}^{n-1}(n-k)^{2} - \sum_{k=0}^{n-1}(n-k) + \sum_{k=0}^{n-1}k^{2} + \sum_{k=0}^{n-1}k\right).
    \end{align*}
    Let $Q_{i,j} = \sum_{m=i}^{j}m^{2}$. Then
    \begin{align*}
        &= \frac{\ell}{2(n-1)}\left(Q_{1,n} - S_{1,n} + Q_{1,n-1} + S_{1,n-1}\right) \\
        &= \frac{\ell}{2(n-1)}\left(2Q_{1,n-1} + n^{2} - n\right).
    \end{align*}
     Using the fact that $Q_{1,m} = \frac{m(m+1)(2m+1)}{6}$ and simplifying:
    \begin{align*}
        &= \frac{\ell}{2(n-1)}\cdot \frac{2n(n - 1)(n+1)}{3} \\
        &= \frac{\ell 2^{\ell}(2^{\ell}+1)}{3}.
    \end{align*} 
    Substituting back into \ref{start-pr-avg} yields:
    \begin{align*}
        \E[p_{r}] &= \frac{1}{\ell 2^{\ell}}\cdot \frac{\ell 2^{\ell}(2^{\ell}+1)}{3} \\
                 &= \frac{2^{\ell} + 1}{3}.
    \end{align*}
\end{proof}

We have verified this value experimentally for $1 \leq \ell \leq 8$. It is interesting to note that the expected asymptotic value differs from the worst-case (maximum) locality by only a constant factor, and grows further apart with $\ell$ from the optimal locality.

Having explored the properties of point locality for single-bit mutations, we now turn our attention to general locality and distance distortion for any variation operator.


\section{Theoretical results on general locality}
\label{sec:theory-general}

General locality is a measurement of the average perturbation of a phenotype for any arbitrary change in genotype, such as the one created by a crossover operator. Again, we start with Rothlauf's definition and refine it for our representation domain. As it turns out, this metric proves to carry no useful information about representations in this domain, because the phenotypic range grows much faster with $\ell$ than the genotypic range.

\subsection{Defining general locality}
\label{subsec:general-definition}

Again we start with Rothlauf's definition, called distance distortion (Eq. \ref{eq:rothlauf-distancedistortion}), and develop an equivalent definition, called \emph{general locality}, that is tailored to our domain. We define the general locality $g_{r}$ for a representation $r$ as the mean difference between phenotypic and genotypic distances between each unique pair of individuals. More formally:

\begin{definition} 
    The general locality for a representation $r$ is
    \begin{equation} \label{general-locality}
        g_{r} \defn \frac{1}{\binom{2^\ell}{2}}\sum_{(s_{1},s_{2}) \in S_{\ell}}|d^{p}(s_{1}, s_{2}) - d^{g}(s_{1}, s_{2})|,
    \end{equation}
    where $S_{\ell}$ is the set of all unordered pairs from $\{0,1\}^{\ell}$ (so $|S_{\ell}| = \binom{2^\ell}{2})$, $d^{p}(s_{1},s_{2})$ is the phenotypic distance $|r(s_{1}) - r(s_{2})|$, and $d^{g}(s_{1},s_{2})$ is the genotypic (Hamming) distance between $s_{1}$ and $s_{2}$. Note that our definition of general locality mirrors Chiam's remoteness preservation and is also equivalent to Rothlauf's $d_{c}$ metric, since $\sum_{i=0}^{n_{p}}\sum_{j=i+1}^{n_{p}}$ is identical to $\sum_{(s_{1},s_{2}) \in S_{\ell}}$.
\end{definition}

We begin our analysis of general locality by proving a lower bound on its value, and continue by proving the asymptotic equivalence of all nonredundant binary integer representations under this metric.

\subsection{Lower bound for general locality}

We compute the tight lower bound on general locality for our domain since it gives an order-of-magnitude estimate for its value, as well as a framework to compute its actual asymptotic value.

\begin{theorem} (Lower bound) \label{dd-lowerbound}
    $g_{r} \geq \frac{1}{\binom{2^\ell}{2}}(\frac{1}{6}(2^{\ell}-1)(2^{\ell})(2^{\ell}+1) - {\ell}2^{2(\ell-1)})$.
\end{theorem}
\begin{proof}
    By definition, $g_{r} = \frac{1}{\binom{2^\ell}{2}}\sum_{(s_{1},s_{2}) \in S_{\ell}}|d^{p}(s_{1}, s_{2}) - d^{g}(s_{1}, s_{2})|$. By the triangle inequality, we have 
    \begin{align*}
        g_{r} &\geq \frac{1}{\binom{2^\ell}{2}}\sum_{(s_{1},s_{2}) \in S_{\ell}}\left(d^{p}(s_{1}, s_{2}) - d^{g}(s_{1}, s_{2})\right) \\
        &= \frac{1}{\binom{2^\ell}{2}}\left(\sum_{(s_{1},s_{2}) \in S_{\ell}}d^{p}(s_{1}, s_{2}) - \sum_{(s_{1},s_{2}) \in S_{\ell}}d^{g}(s_{1}, s_{2})\right) \\
        &= \frac{1}{\binom{2^\ell}{2}}(P - G),
    \end{align*}
    where we let $P = \sum_{(s_{1},s_{2}) \in S_{\ell}}d^{p}(s_{1}, s_{2})$ and $G = \sum_{(s_{1},s_{2}) \in S_{\ell}}d^{g}(s_{1}, s_{2})$. Since $P$ only deals with phenotypes in $\N$, it is equivalent to
    \begin{align*}
        P &= \sum_{i=0}^{2^{\ell} - 1}\sum_{j=i+1}^{2^{\ell} - 1}\left(j-i\right). 
    \end{align*}
    Let the outer sum fix $i$. The inner sum computes the sum of numbers from $1$ to $2^{\ell}-1-i$. This reduces $P$ to
    \begin{align*}
        &= \sum_{i=1}^{n}i + \sum_{i=1}^{n-1}i + \cdots + \sum_{i = 1}^{1}i.
    \end{align*}
    where we let $n = 2^{\ell} -1$ for simplicity.  
    Using the facts that the sum of the first $m$ natural numbers is $\frac{1}{2}m(m+1)$ and the sum of the first $m$ squares is $\frac{1}{6}m(m+1)(2m+1)$, we have
    \begin{align*}
        &= \frac{1}{2}\sum_{i = 1}^{n}i(i+1) \\
        &= \frac{1}{2}\left(\sum_{i = 1}^{n}i^{2} + \sum_{i=1}^{n}i\right) \\
        &= \frac{1}{2}\left( \frac{1}{6}n(n+1)(2n+1) + \frac{1}{2}n(n+1) \right) \\
        &= \frac{1}{6}n(n+1)(n+2).
    \end{align*}
    Substituting $n = 2^{\ell} - 1$ back into the equation yields: 
    \begin{align*}
        &= \frac{1}{6}(2^{\ell}-1)(2^{\ell})(2^{\ell}+1). 
    \end{align*}
    
    Now, since $G$ is the sum of the Hamming distances between all unique pairs of bitstrings, it is equivalent to
    \begin{align*}
        G &= \frac{1}{2}2^{\ell}\sum_{i=1}^{\ell}i\binom{\ell}{i},
    \end{align*}
    because for each of the $2^\ell$ bitstrings, a bitstring has $\binom{\ell}{i}$ other bitstrings with Hamming distance $i$ (choose $i$ of the $\ell$ bits to be flipped). We divide by two because we count each pair twice. Simplifying $G$ gives us
    \begin{align*}
        &= 2^{\ell-1}\sum_{i=1}^{\ell}i\cdot \frac{\ell}{i}\binom{\ell-1}{i-1} \\
        &= {\ell}2^{\ell-1}\sum_{i=1}^{\ell}\binom{\ell-1}{i-1} \\
        &= {\ell}2^{\ell-1}2^{\ell-1} \\
        &= {\ell}2^{2(\ell-1)}.
    \end{align*}
    Substituting $P$ and $G$ back into $g_r$ yields 
    \begin{align*}
        g_{r} &\geq \frac{1}{\binom{2^{\ell}}{2}}(P - G) \\
                &= \frac{1}{\binom{2^{\ell}}{2}}\left(\frac{1}{6}(2^{\ell}-1)(2^{\ell})(2^{\ell}+1) - {\ell}2^{2(\ell-1)}\right).
    \end{align*}
\end{proof}

\subsection{Asymptotic value of general locality}

Here, we prove that the asymptotic value of general locality in our domain is invariant of the actual representation.

\begin{theorem} \label{dd-useless}
    $ g_{r} \sim \frac{1}{\binom{2^{\ell}}{2}}(\frac{1}{6}(2^{\ell}-1)(2^{\ell})(2^{\ell}+1) - {\ell}2^{2(\ell-1)})$ for any representation $r$ on $\ell$ bits. That is, as $\ell$ grows, the value of $g_r$ is independent of the actual representation.
\end{theorem}
\begin{proof}
    The key intuition behind this proof is that for nonredundant binary-integer representations, as $\ell$ grows, the phenotypic distances grow at an asymptotically greater rate than the genotypic distances. We can separate the phenotypic distances from the genotypic distances by partitioning $S_{\ell}$ into two sets $S_{\ell} = S^{p}_{\ell} \sqcup S^{g}_{\ell}$, where $\sqcup$ denotes disjoint union: 
    \begin{align*}
        S^{p}_{\ell} &= \{(s_{1},s_{2}) \in S_{\ell} | d^{p}(s_{1},s_{2}) > d^{g}(s_{1},s_{2})\}, \\
        S^{g}_{\ell} &= \{(s_{1},s_{2}) \in S_{\ell} | d^{p}(s_{1},s_{2}) \leq d^{g}(s_{1},s_{2})\}.
    \end{align*}
    In other words, $S^{p}_{\ell}$ contains all the pairs in $S_{\ell}$ where the two bitstrings have greater phenotypic (Euclidean) distance than genotypic (Hamming) distance, and $S^{g}_{\ell}$ contains all pairs where the two bitstrings have greater or equal genotypic distance than phenotypic distance. We can rewrite $g_{r}$ as (letting $C = 1/\binom{2^{\ell}}{2})$
    \begin{align*}
        g_{r} &= \frac{1}{\binom{2^{\ell}}{2}}\sum_{(s_{1},s_{2}) \in S_{\ell}}|d^{p}(s_{1}, s_{2}) - d^{g}(s_{1}, s_{2})| \\
           &=  C(P(\ell) + G(\ell)),
    \end{align*}
    where we let 
    \begin{align}
        P(\ell) &= \sum_{(s_{1},s_{2}) \in S^{p}_{\ell}}(d^{p}(s_{1},s_{2}) - d^{g}(s_{1},s_{2})), \label{Pl-summation}\\
        G(\ell) &= \sum_{\mathclap{(s_{1},s_{2}) \in S^{g}_{\ell}}}(d^{g}(s_{1},s_{2}) - d^{p}(s_{1},s_{2})).
    \end{align}
    Note that $|S^{g}_{\ell}| \leq \frac{1}{2}{(2\ell - 1)}2^{\ell}$ since each of the $2^{\ell}$ bitstrings can have at most $2\ell - 1$ bitstrings for which their genotypic distance is greater or equal to their phenotypic distance. This is because for an integer $i \geq \ell$, the integers between $i - \ell$ and $i + \ell$ can be represented by bitstrings with genotypic distances greater than phenotypic distances $j \leq \ell$, where we subtract 1 due to the fact that each bitstring only has one other bitstring with Hamming distance $\ell$. We then divide by two since we count each pair twice. Thus $|S^{g}_{\ell}| = \mathcal{O}({\ell}2^{\ell})$. We can now say $|S^{p}_{\ell}| = |S_{\ell}| - |S^{g}_{\ell}| \geq \binom{2^{\ell}}{2} - \frac{1}{2}{(2\ell - 1)}2^{\ell} = \frac{2^{\ell}(2^{\ell}-2\ell)}{2}$, so $|S^{p}_{\ell}| = \Omega (2^{2\ell})$. 
    Consider 
    \begin{align*}
        \lim_{\ell \rightarrow \infty} \frac{|S_{\ell}|}{|S^{p}_{\ell}|} &= \lim_{\ell \rightarrow \infty} \frac{|S^{p}_{\ell}| + |S^{g}_{\ell}|}{|S^{p}_{\ell}|}\\ 
        &= \lim_{\ell \rightarrow \infty}\left(1 + \frac{|S^{g}_{\ell}|}{|S^{p}_{\ell}|}\right) \\
        &= 1 + 0 = 1,
    \end{align*}
    since $|S^{p}_{\ell}| = \Omega (2^{2\ell})$ grows faster than $|S^{g}_{\ell}| = \mathcal{O}({\ell}2^{\ell})$. Thus $|S^{p}_{\ell}|$ dominates $|S^{g}_{\ell}|$, and $|S_{\ell}| \sim |S^{p}_{\ell}|$.
    \\ 
    \\
    We can now perform a similar analysis for $P(\ell)$ and $P(\ell) + G(\ell)$. Note that $G(\ell) \leq |S^{g}_{\ell}|(\ell-1) \leq \frac{1}{2}{\ell}^{2}2^{\ell} - \frac{1}{2}{\ell}2^{\ell}$ since any pair in $S^{g}_{\ell}$ can have a maximum $d^{g} - d^{p}$ of $\ell-1$. Thus $G(\ell) = \mathcal{O}(\ell^{2}2^{\ell})$. Also note that $P(\ell) \geq |S^{p}_{\ell}| \geq \frac{2^{\ell}(2^{\ell}-2\ell)}{2}$ since any pair in $S^{p}_{\ell}$ can have a minimum $d^{p} - d^{g}$ of 1. Thus $P(\ell) = \Omega (2^{2\ell})$. Consider 
    \begin{align*}
        \lim_{\ell \rightarrow \infty}\frac{P(\ell) + G(\ell)}{P(\ell)} &= \lim_{\ell \rightarrow \infty}1 + \frac{ G(\ell)}{P(\ell)}\\
        &= 1 + 0 = 1,
    \end{align*}
    since $P(\ell) = \Omega (2^{2\ell})$ grows faster than $G(\ell) = \mathcal{O} (\ell^{2}2^{\ell})$, and so  $P(\ell) + G(\ell) \sim P(\ell)$. Now we can make a statement about $g_{r}$. We have 
    \begin{align*}
        \frac{1}{C}g_{r} &= P(\ell) + G(\ell) \\
        \lim_{\ell \rightarrow \infty}\frac{g_{r}}{CP(\ell)} &= \lim_{\ell \rightarrow \infty} \frac{P(\ell) + G(\ell)}{P(\ell)}  \\
        \lim_{\ell \rightarrow \infty}\frac{g_{r}}{CP(\ell)} &= 1. 
    \end{align*}
   Thus $\frac{g_{r}}{C} \sim P(\ell)$. Since $|S^{p}_{\ell}| \sim |S_{\ell}|$ and $CP(\ell) \sim g_{r}$, we replace $S^{p}_{\ell}$ with $S_{\ell}$ in Eq. \ref{Pl-summation} to obtain (recall $C = 1/\binom{2^{\ell}}{2}$)
    \begin{align*}
        g_{r} &\sim \frac{1}{\binom{2^{\ell}}{2}}\sum_{(s_{1},s_{2}) \in S_{\ell}}\left(d^{p}(s_{1},s_{2}) - d^{g}(s_{1},s_{2})\right) \\
               &= \frac{1}{\binom{2^{\ell}}{2}}\left(\frac{1}{6}(2^{\ell}-1)(2^{\ell})(2^{\ell}+1) - \ell 2^{2(\ell-1)}\right),
    \end{align*}    
    which was found in the proof of Theorem \ref{dd-lowerbound}. Thus $g_{r}$ and $\frac{1}{\binom{2^\ell}{2}}(\frac{1}{6}(2^{\ell}-1)(2^{\ell})(2^{\ell}+1) - {\ell}2^{2(\ell-1)})$ are asymptotically equal for any representation $r$. 
\end{proof}

Theorem \ref{dd-useless} states that all representations have asymptotically equal general locality. Since SB and BRG are well-defined for any number of bits, it follows that their general localities are asymptotically equal as well. In fact, on just their 11-bit representations (with only $n_p=2,048$ possible solutions, where a brute-force search likely outperforms many GEAs), we calculate $g_{SB} = 677.497$ and $g_{BRG} = 677.502$, or only a $0.0007\%$ difference. The implication is therefore that general locality, and by extension, the original definition of distance distortion, are not useful as a description or prediction of any representation in this domain. In a domain where phenotypic and genotypic distances grow with $\ell$ at similar rates, such as with unary representation, these definitions may have more power.

\section{Experimental results}
\label{sec:experiments}

As the previous section proved, Gray encoding exhibits no general or point locality advantage over standard binary. On the other hand, several studies found a performance advantage for Gray under various GEAs~\citep{whitley97:representation, whitley97:search, whitley99:free, mathias94:transforming}.
As a first step towards understanding why stronger locality does not always lead to a GEA performance advantage, this section expands on past experimental results and analyzes the factors that lead to better performance. 
Our experiments progress from the simple to the complex, to allow a tractable analysis of the effect of point locality, as well as an evaluation of both localities in richer and more realistic scenarios.
All of our source code, representations, choices of parameters, and Markov models can be found in the supplementary material and at \url{https://github.com/shastrihm/GAMO-R}.

\subsection{Simulated annealing}
\label{subsec:SA}

One very simple GEA is single-organism simulated annealing (SA), which we apply to the general \mbox{ONEMAX} problem to replicate Rothlauf's results in Sec. 5.4 of his book~\citep{rothlauf06:representations}. To briefly recap his general ONEMAX experiment, Rothlauf defines an $\ell$-bit genotype $x$ that is translated to an integer phenotype $x^p \in [0:2^{\ell}-1]$ by a given representation. The fitness of the phenotype is evaluated against a target $a$ with the function
\begin{align*}
    f_{p}(x^{p}) &= x_{max} - |x^{p} - a|, ~~~ \text{(ibid. Eq. 5.2)} 
\end{align*}
where $x_{max}$ is defined as the maximum phenotypic value, $2^{\ell}-1$, and is attained only at the global maximum $a$.
The genotype is iteratively mutated with a random single-bit flip, and the offspring replaces the parent if it improves its fitness, or at a probability determined by a Boltzmann process. This probability decreases both with the fitness difference between the parent and offspring, and with a global temperature parameter that decreases every iteration.

As in the original study, we set $\ell=5$, the initial temperature to $50$, and the cooling factor to $0.995$, and experiment with different representations and $a$ values. Each experiment is run for a few thousand generations (mutations) until it converges on a solution. Finally, we repeat each experiment concurrently and independently for thousands of different random starting genotypes and record for each generation the percentage of experiments (genotypes) that are at the global optimum.\footnote{The book concatenates ten 5-bit genotypes to one bitstring and sums up their individual fitness values. To simplify the analysis, we use a single 5-bit organism at a time. Our code allows an exact reproduction of Rothlauf's result if concatenation is selected.}

\begin{figure*}[htp]
    \centering
    \includegraphics[width=0.47\textwidth]{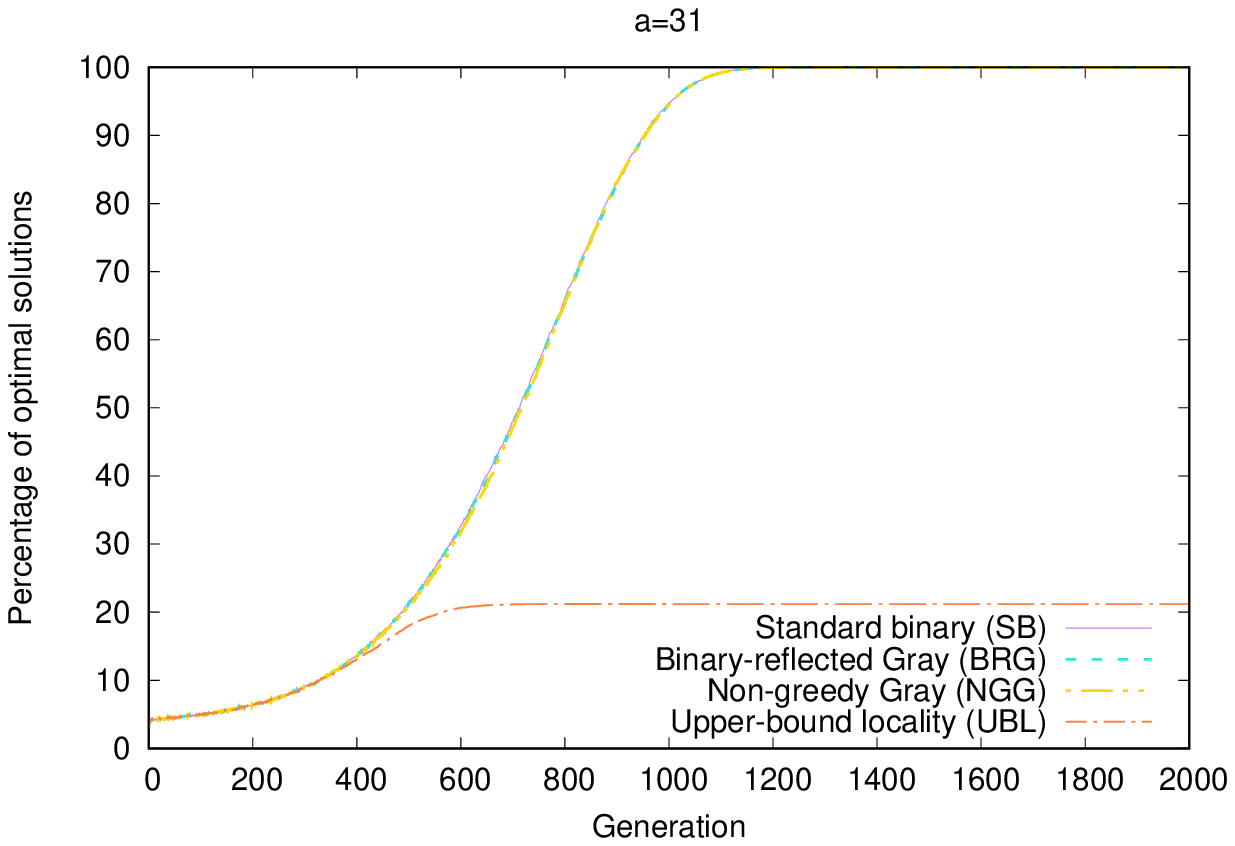}
    \includegraphics[width=0.47\textwidth]{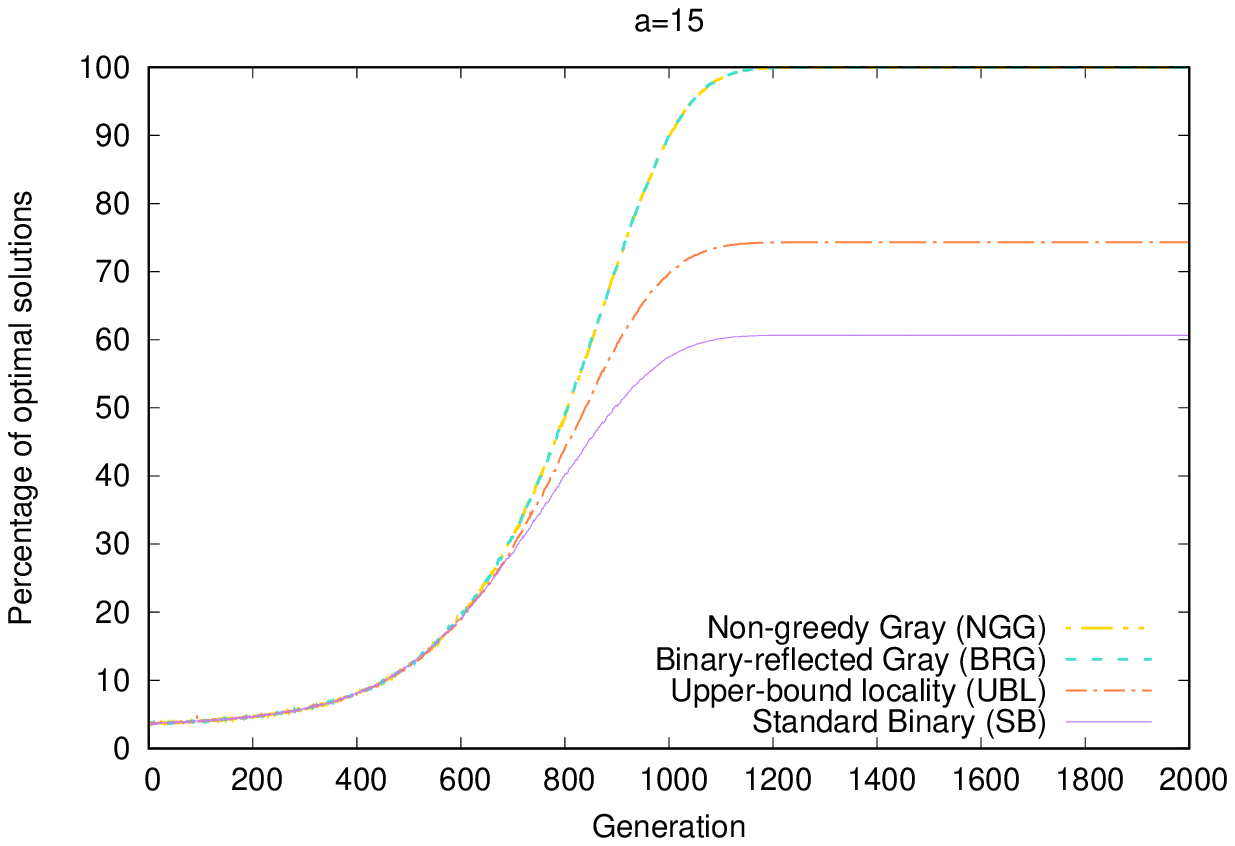}
    \caption{Performance of four representations under simulated annealing for a 5-bit ONEMAX problem averaged over 100,000 random seeds. Higher results are better.}
    \label{fig:SA31-15}
\end{figure*}

Rothlauf compared the GEA's performance across two representations, SB and BRG, using either $a=31$ (where both representations perform equally) and $a=15$, where SB often gets trapped on the local maximum phenotype 16 and thus significantly underperforms BRG. Our code produced the same results, as shown in Fig.~\ref{fig:SA31-15}.

In the original study, Rothlauf summarizes the worse performance of SB as follows: ``The binary encoding has problems associated with the Hamming cliff and has low locality. Due its low locality, small changes of the genotype do not always result in small changes of the corresponding phenotype'' (p. 136). However, as the previous section showed, BRG has no locality advantage over SB by either metric. In particular, the locality metric that captures the mean degree of variation across single-bit mutations, namely point locality, is equal for both: $p_{SB}=p_{BRG}=\frac{2^\ell - 1}{\ell}=6.2$. Weak locality cannot therefore be explain of performance differences in this case.

The other explanation about the Hamming cliff requires some unpacking. A Hamming cliff, which we loosely define as a genotype whose immediate genotypic neighbors are not also phenotypic neighbors, can represent a local phenotype maximum~\citep{rowe04:evolution}. That is because any genotype other than the global maximum, whose immediate genotypic neighbors (i.e., one bit-flip away) are not phenotypic neighbors, will have nonlocal jumps in values of the fitness function. These jumps could therefore all be farther away from the global maximum's fitness. Such genotypes are disfavored by the selection process to mutate, because any mutation lowers the fitness value. With a certain combination of temperature, cooling factor, and fitness parameters, the Boltzmann selection operator can develop a nonzero probability of getting stuck in the suboptimal genotype, leading to underperformance of the GEA. This is exactly what happens in the example of $a=15$, where BRG has only one maximum at 15, but SB also has a suboptimal local maximum at 16, where no single-bit mutation improves fitness. Over time, SB has $\approx{0.4}$ probability of getting stuck in this local optimum, leading to $\approx{60\%}$ optimal solutions over all random seeds.

Gray encoding, on the other hand, is guaranteed not to have more than one optimum, because every non-optimal phenotype has at least one genotypic neighbor---one bit flip away---that is also a phenotypic neighbor---one integer away---in the right direction. Whenever such a mutation happens, it is selected by the Boltzmann process, converging eventually to the single optimum. This property is invariant of the point locality value. To demonstrate this, we synthesized a Gray representation based on Claim~\ref{nongreedygray} and termed it non-greedy Gray (NGG).\footnote{NGG = [0, 1, 19, 2, 31, 28, 20, 3, 23, 26, 24, 25, 22, 27, 21, 4, 13, 14, 18, 15, 30, 29, 17, 16,12, 9, 11, 10, 7, 8, 6, 5].} By construction, for $\ell = 5$, $p_{NGG} = 8.35 > p_{SB}$. Nevertheless, NGG performs perfectly for both target values, as Fig.~\ref{fig:SA31-15} shows.

To take this point to the extreme, we synthesized a worst-case representation using the construction in Claim~\ref{worst-rep} and termed it upper-bound locality (UBL).\footnote{UBL = [24, 1, 4, 19, 15, 16, 21, 13, 9, 26, 18, 0, 23, 12, 6, 22, 3, 28, 20, 14, 30, 7, 5, 27, 29, 10, 8, 31, 2, 17, 25, 11].} By construction, $p_{UBL} = 2^{4} = 16 > p_{SB}$. And although UBL performs poorly for $a=31$ and suboptimally for $a=15$, it still outperforms SB in the latter case.

The two point-locality extremes, SB and UBL, outperform each other for different target values. Even among representations with the same locality, performance is uneven. We must therefore conclude again that locality does not determine performance for this particular problem. The second explanation, the existence of Hamming cliffs, is also insufficient to predict performance~\citep{chakraborty03:analysis}. 

Gray coding is sometimes described as higher-performing than SB because it can produce fewer local maxima for many useful problems~\citep{whitley97:representation}.
But the local maxima count on its own is not a strong predictor of GEA performance for this problem, because the probability of getting stuck in any specific maximum depends on its fitness distance from its genotypic neighbors.
For example, UBL has four Hamming cliffs (local maxima) with $a=15$, compared to SB's two, and yet UBL still outperforms SB. In another example with $a=29$, SB has three local maxima, and yet it still converges to the global optimum every single time. 

In summary, neither locality nor local maxima count can predict performance for the general ONEMAX problem under Boltzmann selection. This holds true for many other functions as well~\citep{chakraborty03:analysis}.
A more nuanced explanation of the GEA performance arises from two observations: that the linear fitness function penalizes local maxima that are farther from the global maximum (in phenotypic distance), and that the Boltzmann selection process gives preference to mutations that minimize phenotypic distance.

The interaction between these two factors can be modeled with a simple Markov chain, which fully predicts the probability of converging to the local maximum in the absence of a cooling factor.
Adding the cooling factor complicates the model, but it can still be computed with a dynamic probability function. The main constraint this GEA imposes on some representations is the inability to break free from some local maxima. In the next section we turn to a different GEA, so that we may evaluate the effect of point locality with different mutation and selection operators that can always jump out of a local maximum.


\subsection{Evolutionary strategies}
\label{subsec:ES}

Our next experiment introduces two changes to the previous GEA. First, we use simple elitist selection instead of Boltzmann selection. An offspring only replaces a parent if it improves its fitness. This change would make local maxima impassable for single-bit mutation, so we change the mutation operator as well. Mutations can now flip any bit independently with probability $m$, so that any genotype can be mutated to any other genotype at some positive probability that depends on $m$ and the Hamming distance between them. Therefore any genotype could in principle be mutated to the global maximum in a single step. From a locality point of view, invoking multiple bit flips per generation has the same effect as invoking a single random bit flip for multiple generations, so a representation with strong point locality under a single bit flip should also exhibit a smaller mean change in phenotype value in this experiment.

This modified GEA is a type of \emph{Evolutionary Strategies} on discretized numbers, denoted as (1+1)-ES.
\cond{Mühlenbein \citep{muhlenbein92:genetic}}{\citet{muhlenbein92:genetic}}
investigated this GEA in the context of the same ONEMAX problem by developing an analytical model for the expected number of generations to reach the optimum solution. Using a simpler fitness function that merely counts the number of `1' bits in the genotype, his model predicts the mean number of generations to converge on the optimum when the mutation rate is $m = \frac{1}{\ell}$. His prediction holds for any representation since the fitness function ignores the phenotype. Indeed, when we ran the same simulation on various $\ell$ values larger than 5, we get similar empirical convergence times as M{\"u}hlenheim's experimental results, irrespective of the representation.

Reintroducing Rothlauf's generalized ONEMAX fitness function adds back a dependency on the representation, as Table~\ref{tab:ES} shows. Modeling this performance with a Markov chain confirms the empirical results and the differences between the representations. This is an example where SB actually outperforms both Gray encodings, at least for $a=31$. In this case, note that a mutation improves fitness if and only if it flips a bit to `1'. Improving mutations, which are the only ones that can change the genotype, are independent of each other and of any sequence, so the order of `0' to `1' bit flips does not matter. This leads to swift convergence to the optimum. Conversely, in all other representations, some sequences of fitness improvement require first flipping a specific bit one way, then later the other way. It imposes an ordering or interdependency of mutations that prolongs convergence. It is especially bad for UBL, because many fitness improvements require more than one bit flip at the same time, which occurs less frequently.

When we switch to $a=15$, we introduce the same handicap to SB, because now it has to flip every single bit at once when trying to escape the phenotype 16, at a low probability of $m^{\ell}$. Despite the dismal performance, it is worth noting that SB still performs better under (1+1)-ES than under simulated annealing (SA): in 2,000 generations, it reaches the optimal solution in $65.9\%$ of the runs, compared to SA's $60.6\%$, and if allowed to evolve for more generations, all runs will eventually find the optimum.

It is also worth noting that NGG slightly underperforms BRG for both target values, which could be a related to its weaker locality. But overall, locality is a poor explanation for (1+1)-ES performance in this experiment, because SB both overperforms BRG with $a=31$ and underperforms it with $a=15$, despite both representations having the exact same point locality.

\begin{table}[h]
    \caption{Mean generations to reach optimum in a 5-bit (1+1)-ES generalized ONEMAX evaluation with a mutation rate of 0.2, averaged across 100,000 runs of up to 1,000 generations each. Runs that did not reach the optimal solution were excluded from the mean.  Lower results are better.}
    \label{tab:ES}
    \centering
    \begin{tabular}{c|c|c|c|c}
    \hline
         Optimum (a) & SB & BRG & NGG & UBL \\
    \hline
    \hline
         31 & 17.5 & 23.8 & 28.3 & 184.8 \\
    \hline
         15 & 103.6 & 21.3 & 23.3 & 124.8 \\
    \hline
    \end{tabular}
\end{table}


\subsection{Genetic algorithms}
\label{subsec:GA}

For our last experiment, we add three more layers of complexity. First, we add a population element with fitness-proportionate selection (roulette-wheel).\footnote{Tournament selection yielded substantively identical results.} Second, we add a recombination operator, specifically single-point crossover. And finally, we expand our evaluation functions to De~Jong's test suite, which can be more challenging to solve with a GEA than ONEMAX~\citep{dejong75:phdthesis}. These five functions take real numbers as inputs, but De~Jong and subsequent studies used fixed-point numbers as inputs, which lie in the same binary-to-integer scope as the rest of this paper.

To compare different representations under a genetic algorithm (GA), we attempted to replicate the experiments of \cond{Caruana and Schaffer \citep{caruana88:representation}}{\citet{caruana88:representation}}, which in turn derive their GA parameters from \cond{Grefenstette \citep{grefenstette86:optimization}}{\citet{grefenstette86:optimization}}.
We measured the online performance under SB, BRG, NGG, and UBL representations, where online performance is defined as the average fitness of all function evaluations to the current point.
Their results and ours are shown in Table~\ref{tab:crossover}.

\cond{}{\setlength{\tabcolsep}{2pt}}
\begin{table*}[h]
\renewcommand{\arraystretch}{1.3}
\caption{GA online performance for four representations averaged over 3,000 runs. 
\cond{Caruana and Schaffer \citep{caruana88:representation}}{\citet{caruana88:representation}} results averaged over 5 runs in parentheses. Lower values denote better performance.}
\label{tab:crossover}
\centering
\tabcolsep=0.09cm
\begin{tabular}{c|c|c|c|c|c|c|c}
    \hline
    Function  &  Description & Dimensions & Optimum & SB & BRG & NGG & UBL  \\
    \hline
    \hline
    f1   & Parabola & 3 & 0 & 2.49 (1.66) & 2.03 (1.43) & 2.03 & 6.70 \\
    \hline
    f2    & Rosenbrock's Saddle & 2 & 0 & 32.56 (25.16) & 22.31 (13.18) & 22.35 & 199.48\\
    \hline
    f3    & Step function & 5 & -30 & -26.48 (-27.78) & -25.10 (-27.16) & -25.06 & -17.86\\
    \hline
    f4    & Quadratic with noise & 30 & 0 & 41.46 (24.28) & 32.32 (21.77) & 32.28 & 65.83 \\
    \hline
    f5    & Shekel's foxholes & 2 & $\approx 1$ & 56.71 (30.78) & 35.86 (20.68) & 35.84 & 149.92\\
    \hline
\end{tabular}
\end{table*}

Unfortunately, we could not uncover the full details or source code of their implementation, and had to construct our own experiment from scratch, likely with some different parameter or implementation choices. These differences could help explain why our results are not identical to the original study's. Another explanation could be that the original study averaged each experiment over only five trials, which we found too noisy. Instead, we averaged performance over 3,000 trials (90,000 fitness evaluations) for each experiment to increase statistical robustness.

While our results differ in quantity, they agree in quality. Caruana and Schaffer found BRG to perform similar to or better than SB on the minimization of all five test functions, as have we and others~\citep{hinterding95:nature}. While Caruana and Schaffer recognize that an encoding is as likely to outperform another encoding on an arbitrary search, they believe that ``many common functions which have ordered domains have local correlations between domain and range. A GA using Gray coding will often perform better on this class of functions than one using binary coding because the Gray coding preserves this structure better than binary code.'' The authors suggest that it is the introduction of Hamming cliffs that biases SB away from preserving the domain structure. The fact that the weaker-locality NGG performs so closely to BRG supports this explanation.

Their paper does not clarify whether and how the crossover mechanism interacts with the representation. For example, there is evidence that Gray encoding interferes with standard crossover operators because it disrupts common schemata~\citep{weicker10:binary}, and that if crossover is the only variation operator, SB can actually outperform BRG on ONEMAX~\citep{rothlauf02:binary}.

What \emph{is} clear from our experiment is that these behaviors are not captured by the locality metrics. BRG outperforms SB despite having the same point locality, and is equal in performance to NGG, with the weaker locality. Additionally, all four representations share the same values for the distance distortion, general locality, or remoteness preservation metrics. 

To summarize all three experiments, none of our results shows that strong locality leads necessarily to better GEA performance. The next section suggests some reasons for this counterintuitive result.

\section{Discussion}
\label{sec:discussion}

\subsection{Locality and GEA performance}
\label{subsec:eventual}

Why should strong locality improve GEA performance? We can list two reasons. 
First, strong locality helps preserve building blocks (schemata) across variation operators, because it supports the linkage between genotypic and phenotypic building blocks~\citep{vose91:generalizing}.
Second, specifically for mutation operators, strong locality enables localized, hill-climbing, or gradient search, by effecting small changes to the genotype, such as single-bit mutation. With strong locality, small genotypic changes lead to small phenotypic changes, and because many practical fitness functions are locally continuous, they also lead to small fitness changes. In Rothlauf's words~\citep{rothlauf03:locality}: 
 
 \begin{quote}
 ...low-locality representations randomize the search process and make problems that are easy for mutation-based search more difficult and difficult problems more easy.  Although low-locality representations increase the performance of local search on difficult, deceptive problems this is not relevant for real-world problems as we assume that most problems in the real-world are easy for mutation-based search.
\end{quote}

Despite these reasons, we found no strong association between various locality measures and GEA performance in our work and others'. 
One possible explanation is that the common test functions we evaluated are better-suited for Gray encoding, perhaps because of their sensitivity to Hamming cliffs~\citep{caruana88:representation, whitley99:free}. Another explanation is that in the domain we study here, nonredundant binary-integer mappings, our locality and distance metrics are skewed: phenotypical distances grow exponentially with $\ell$, while genotypic (Hamming) distances grow linearly. If we use instead the redundant unary bitstring-to-integer representation, for example, then phenotypic distances also grow linearly with genotypic distances, but the GEA would typically underperform nonredundant representations~\citep{rothlauf06:representations}. As another example, in the separate domain of program trees, some definitions of locality have been found to correlate well with  GEA performance~\citep{galvan11:defining}. 

Regardless, even if in our domain locality does not affect eventual performance, we think it has a role in predicting GEA performance over time, especially in regards to the mutation operator.
The role of the mutation operator in GEAs is to introduce genotypical and phenotypical variation~\citep{hinterding95:nature}. In other words, mutations serve to diversify the population so that more potential solutions are evaluated. Locality, as defined here, quantifies the average impact of single mutations, with strong locality leading to smaller overall phenotypical variation. Whether or not strong locality improves GEA performance depends on the effect of diversity on performance, which itself varies over time.

\subsection{Phenotypic diversity over time}
\label{subsec:time}

In the process of converging towards an optimal solution, GEAs combine and balance two different search strategies, exploration and exploitation~\citep{crepinsek13:exploration, weicker10:binary}. In the former, high variance or diversity is desired so that many subspaces of the fitness landscape are explored. When the GEA identifies a promising subspace, large variations in phenotype are more likely to be disruptive rather than helpful to fitness, so small perturbations are usually preferred~\citep{rowe04:properties, mathias94:transforming}. If the only variance-introducing operator in a GEA is single-bit mutation, then we should expect strong-locality representations to underperform weak-locality representations in the early phase of the search, and outperform in the latter phase. 

This difference matters most if our GEA is meaningfully constrained by computational resources. The GEA's task then is to quickly explore the solution space for a ``good enough'' solution, rather than an eventual convergence towards an optimum. If a GEA is configured such that the expected fitness is relatively high after a few generations, and before convergence, then it has a higher probability of finding a decent solution with fewer resources, even if it is slower to converge later. This distinction is similar to the one between online performance and offline performance of the GEA~\citep{grefenstette86:optimization}.

As an example, consider a comparison of performance over time between a strongest-possible locality representation and a weakest-possible locality representation in the the simulated annealing experiment (Sec.~\ref{subsec:SA}). We picked two arbitrary target $a$ values for SB and UBL such that both representations have the same number of local maxima, four. Because we are interested in the likelihood of getting a ``good enough'' solution in limited time, rather than an optimal solution, we measure the expected (mean) fitness per generation across 100,000 runs, instead of the percentage of perfect solutions. 

\begin{figure}[htp]
    \centering
    \includegraphics[width=0.47\textwidth]{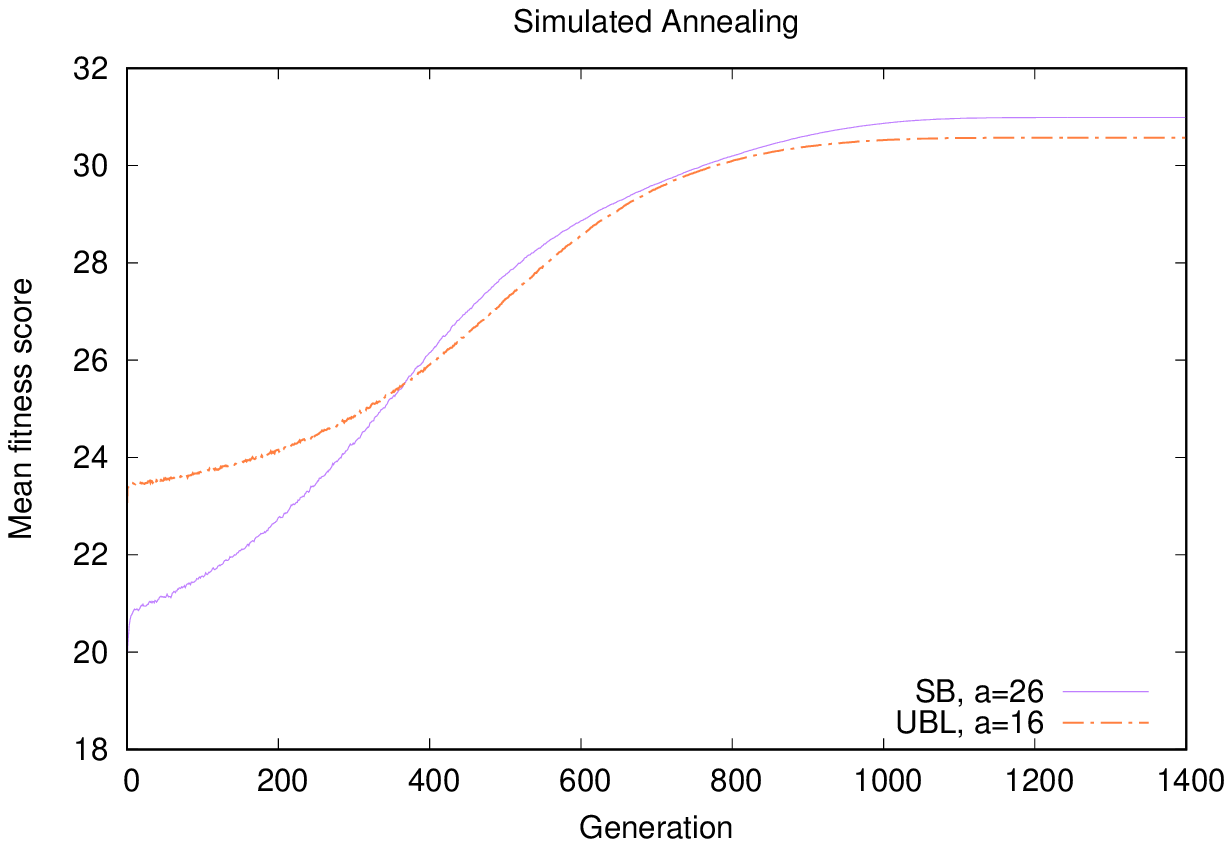}
    \includegraphics[width=0.47\textwidth]{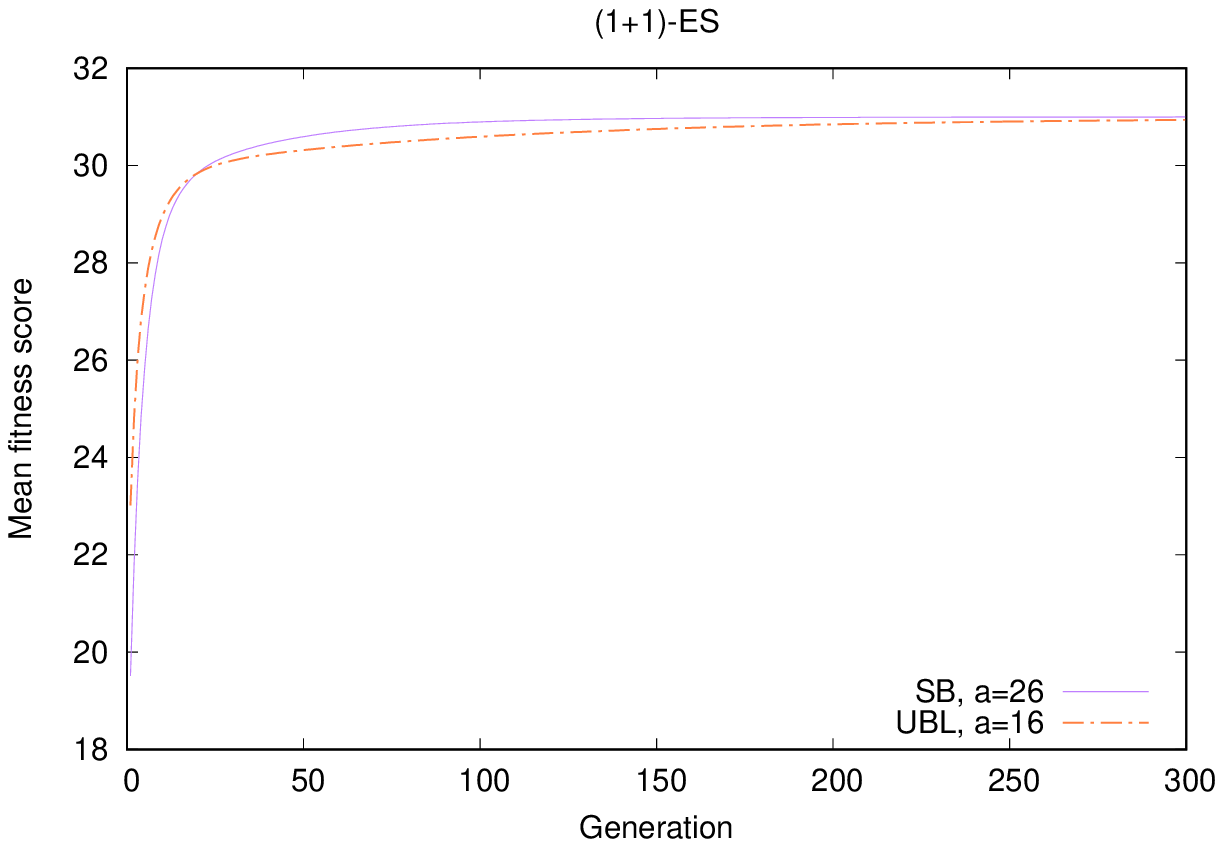}
    \caption{Convergence speed for two different 5-bit general ONEMAX problems with four local maxima each and either simulated annealing or evolutionary strategies. Fitness is averaged over 100,000 trials.}
    \label{fig:SA-comparison}
\end{figure}

Fig.~\ref{fig:SA-comparison} shows that for long-enough runs (starting from about 400 generations), both representations perform about the same, with a small eventual advantage to SB. But in the earlier generations, during the exploration phase, SB significantly underperforms UBL. In other words, this example shows that if we must stop our simulation at an early point before convergence, the representation with the weaker locality is more likely to produce a solution of higher fitness. But after that threshold has been passed, the stronger-locality representation may be more suitable for fine-grained exploitation to locate a higher-fitness solution. On the other hand, sometimes we are specifically interested in local or hill-climbing search~\citep{mitchell94:outperform}, in which case Gray encoding may be preferable to SB, even with the same locality, because it always offers a genotypic neighbor that can move the phenotype towards the maximum~\citep{rowe04:properties}.

\begin{figure*}[htp]
    \centering
    \includegraphics[width=0.45\textwidth]{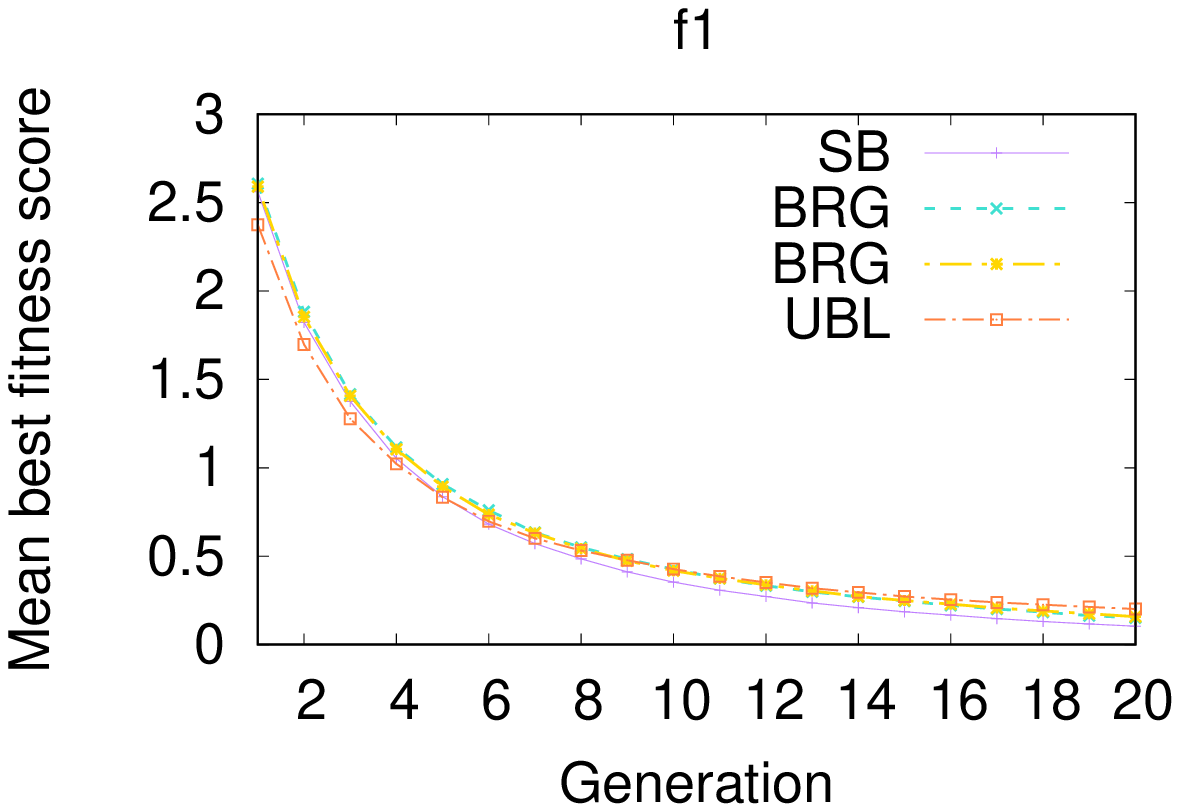}
    \includegraphics[width=0.45\textwidth]{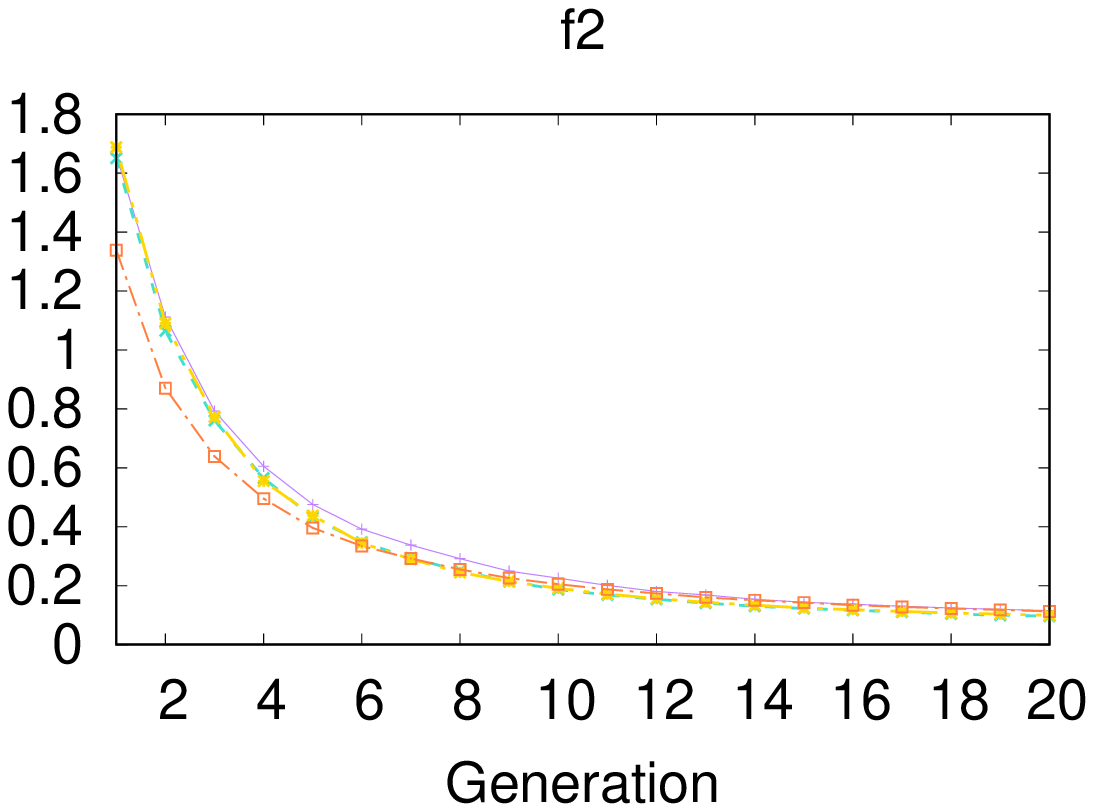}
    \includegraphics[width=0.45\textwidth]{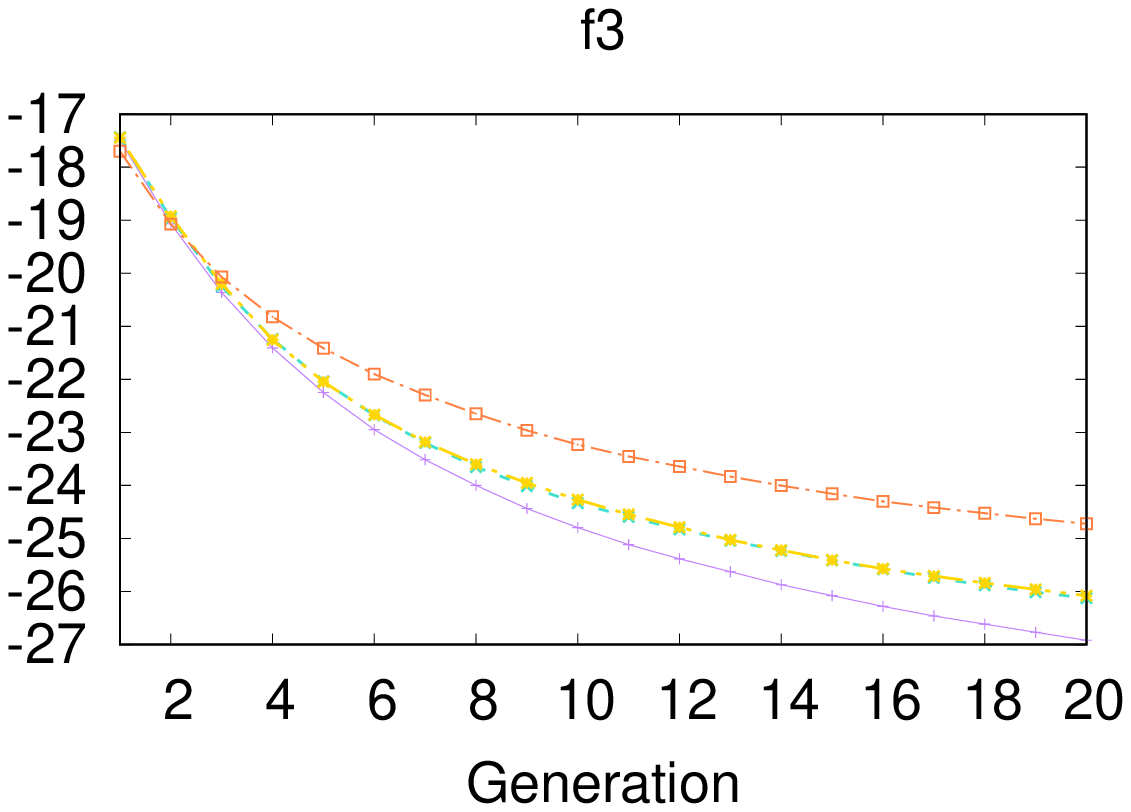}
    \includegraphics[width=0.45\textwidth]{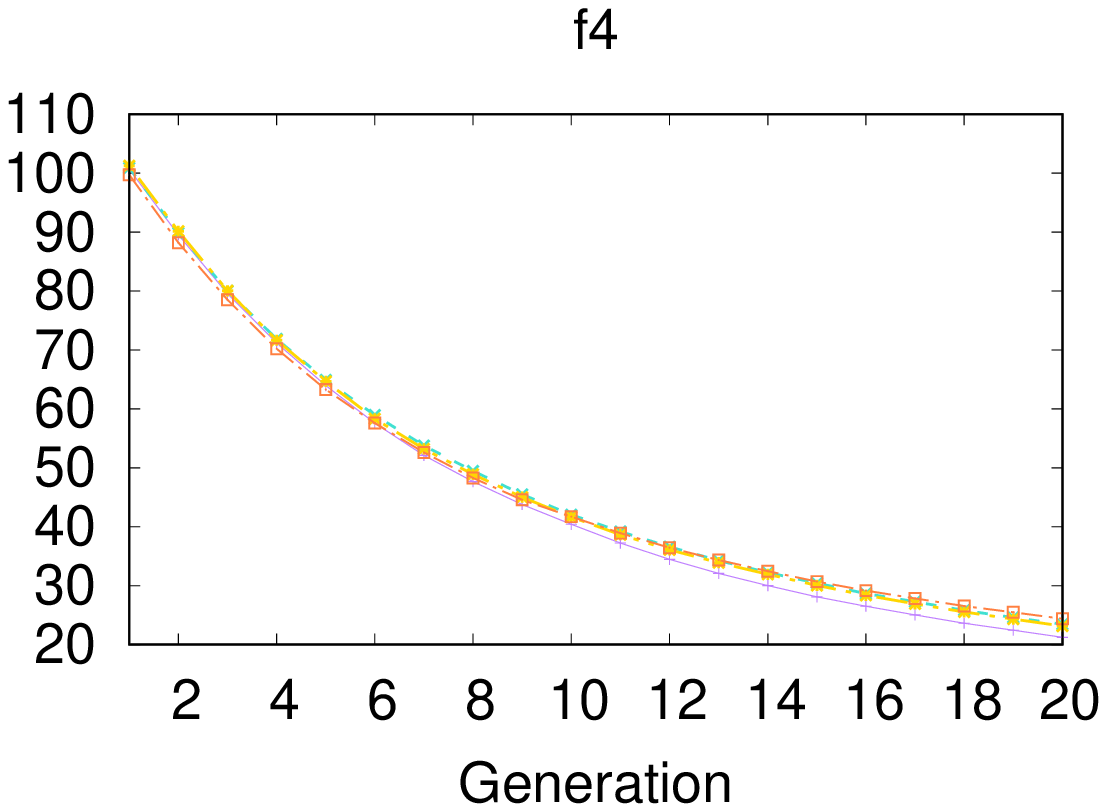}
    \includegraphics[width=0.45\textwidth]{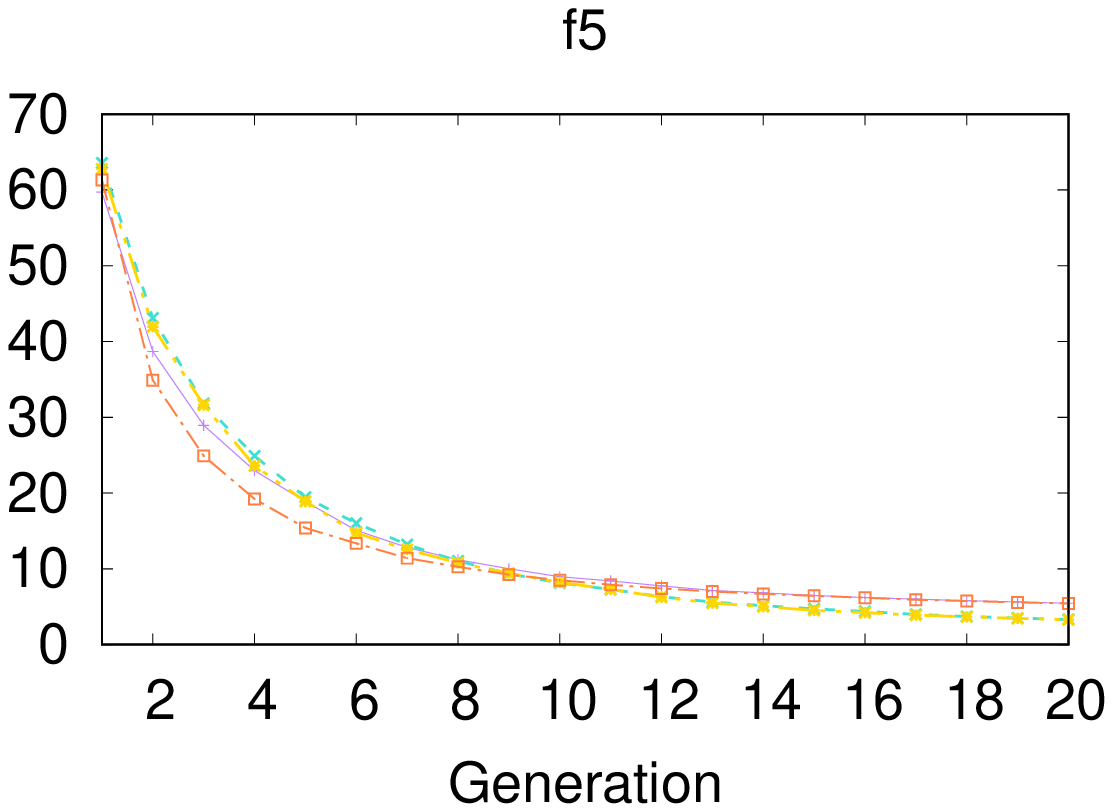}
    \caption{Performance of four representations for the first 20 generations of a genetic algorithm. Each data point averages fitness values from the best solution in that generation across 3,000 trials.}
    \label{fig:ga-over-time}
\end{figure*}

When we repeat the experiment with (1+1)-ES, we see the same effect, where UBL starts out stronger but is then quickly surpassed by SB. In this case, both representations end up with the same eventual performance. Even the GA experiments show the same locality effect (Fig.~\ref{fig:ga-over-time}), where UBL always starts out stronger than the other representations, even if only for a very short time.

This change in the role of mutations over time is also why some GEAs vary the mutation rate or operator over the course of the search~\citep{eiben98:evolutionary, gomez03:adaptive, hansen01:CMA-ES, zhao07:evolutionary}. An alternative approach would start the search with a weak-locality representation, and switch to a strong-locality representation as we move from the exploration phase to the exploitation phase.

Some consider the recombination operator as the main exploration tool, with mutation as the main exploitation tool~\citep{caruana88:representation}, while others disagree~\citep{hinterding95:nature}. We think a more nuanced discussion needs to include the locality of the representation under the chosen mutation operator. Our results suggest that strong-locality representations are more suitable for an exploitation-oriented mutation operator, since they limit the perturbation of the phenotype, and vice-versa.

Changing representation randomly during the GEA execution may not be a practical tool to improve performance, because the expected number of local optima is too high~\citep{whitley97:representation}. As we have shown in Sec.~\ref{subsec:expected_locality}, the expected locality is also too high, closer to the pessimal locality than to the optimal locality, on average. But if our goal for the exploration phase of the GEA is to generate significant genotypic diversity and coverage of the search subspaces, then shifting to and between random representations may help this goal. To the best of our knowledge, no one has tried this approach, and it remains as promising future work for analysis. 

\section{Conclusion and future work}
\label{sec:conclusions}

Various properties of GEA representations have been studied to help explain GEA performance, such as the number of induced local optima or the existence of Hamming cliffs. Among these properties, locality is particularly interesting to analyze, because it is independent of the fitness landscape and can be computed precisely, which is not always the case for other properties.

But if strong locality is also a strong factor in good GEA performance, as has been expressed in the literature, then we do not always have the right metrics to measure it. Our own point and global locality metrics, based on the works of Rothlauf and others, show no clear relationship to offline GEA performance in the domain of nonredundant binary-integer representations. This negative finding may not hold for other domains, and we plan to explore locality metrics specific to redundant, nonbinary, or noninteger representations as well. It also does not hold for resource-constrained GEAs, since for short-enough executions, weak-locality representations do appear to outperform strong-locality representations in our experiments, as they emphasize exploration over exploitation.

Both the $d_m$ and $d_c$ locality metrics (and consequently, $p_r$ and $g_r$) estimate the locality of a representation by using the sum of phenotypic distances, which grows exponentially in our domain. Another path for future research might be to use other statistics as estimators of locality that grow more slowly with the bitstring length, such as the minimum phenotypic distance, maximum, or standard deviation.
In addition, both Rothlauf's locality metric and our own equivalent point locality focus on the single-bit mutation operator. Other locality metrics could look at different operators, and perhaps combine them with a crossover operator to expand on distance distortion.

Of the two most common binary-integer encodings, Gray has often been shown to outperform standard binary on many GEAs and test functions. There are various context-dependent explanations for this advantage, such as the Hamming cliff, differing numbers of local optima, and the properties of the typical test functions. What our study has shown, both analytically and empirically, is that the locality of the representations cannot reliably be one of these explanations. A complete characterization and understanding of this performance difference between binary and Gray remains an interesting research question, and a different locality metric may be able to shed more light.

\subsubsection*{Acknowledgments}

We wish to thank Prof. Dr. Franz Rothlauf for fruitful discussions at the beginning of this project.

\bibliographystyle{plain}
\bibliography{representation}

\end{document}

%% file: bad_gray.tex
\tikzset{every picture/.style={line width=0.75pt}} 

\begin{tikzpicture}[x=0.75pt,y=0.75pt,yscale=-1,xscale=1]

\draw [line width=0.75]    (141.44,90.57) -- (141.44,44.64) ;
\draw [line width=0.75]    (306.95,90.57) -- (306.95,44.64) ;
\draw [line width=0.75]    (472.45,90.57) -- (472.45,44.64) ;
\draw [color={rgb, 255:red, 245; green, 166; blue, 35 }  ,draw opacity=1 ][line width=0.75]    (632.83,90.53) -- (632.83,46.2) ;
\draw [shift={(632.83,44.2)}, rotate = 450] [color={rgb, 255:red, 245; green, 166; blue, 35 }  ,draw opacity=1 ][line width=0.75]    (10.93,-3.29) .. controls (6.95,-1.4) and (3.31,-0.3) .. (0,0) .. controls (3.31,0.3) and (6.95,1.4) .. (10.93,3.29)   ;
\draw [color={rgb, 255:red, 74; green, 144; blue, 226 }  ,draw opacity=1 ][line width=0.75]    (374.6,245.53) -- (374.6,201.2) ;
\draw [shift={(374.6,199.2)}, rotate = 450] [color={rgb, 255:red, 74; green, 144; blue, 226 }  ,draw opacity=1 ][line width=0.75]    (10.93,-3.29) .. controls (6.95,-1.4) and (3.31,-0.3) .. (0,0) .. controls (3.31,0.3) and (6.95,1.4) .. (10.93,3.29)   ;

\draw (38.34,69.9) node    {$100$};
\draw (78.46,31.63) node    {$110$};
\draw (38.34,146.44) node    {$\mathbf{\textcolor[rgb]{0.96,0.65,0.14}{000}\textcolor[rgb]{0.96,0.65,0.14}{^{0}}}$};
\draw (142.64,39.28) node [anchor=south] [inner sep=0.75pt]    {$111$};
\draw (110.55,69.9) node    {$101$};
\draw (78.46,100.52) node    {$010$};
\draw (142.64,99.75) node    {$011$};
\draw (110.55,146.44) node    {$001$};
\draw (203.85,69.9) node    {$100$};
\draw (243.97,31.63) node    {$110$};
\draw (203.85,146.44) node    {$\textcolor[rgb]{0.29,0.56,0.89}{000}\textcolor[rgb]{0.29,0.56,0.89}{^{0}}$};
\draw (308.15,39.28) node [anchor=south] [inner sep=0.75pt]    {$111$};
\draw (276.06,69.9) node    {$101$};
\draw (243.97,100.52) node    {$010$};
\draw (308.15,99.75) node    {$011$};
\draw (276.06,145.67) node    {$\mathbf{\textcolor[rgb]{0.96,0.65,0.14}{001}\textcolor[rgb]{0.96,0.65,0.14}{^{1}}}$};
\draw (369.36,69.9) node    {$100$};
\draw (409.47,31.63) node    {$110$};
\draw (369.36,146.44) node    {$\textcolor[rgb]{0.29,0.56,0.89}{000}\textcolor[rgb]{0.29,0.56,0.89}{^{0}}$};
\draw (473.66,39.28) node [anchor=south] [inner sep=0.75pt]    {$111$};
\draw (441.57,69.9) node    {$101$};
\draw (409.47,100.52) node    {$010$};
\draw (473.66,99.75) node    {$\mathbf{\textcolor[rgb]{0.96,0.65,0.14}{011}\textcolor[rgb]{0.96,0.65,0.14}{^{2}}}$};
\draw (441.57,146.44) node    {$\textcolor[rgb]{0.29,0.56,0.89}{001}\textcolor[rgb]{0.29,0.56,0.89}{^{1}}$};
\draw (529.74,69.68) node    {$100$};
\draw (569.85,31.07) node    {$110$};
\draw (529.74,146.89) node    {$\textcolor[rgb]{0.29,0.56,0.89}{000}\textcolor[rgb]{0.29,0.56,0.89}{^{0}}$};
\draw (634.04,38.83) node [anchor=south] [inner sep=0.75pt]    {$\mathbf{\textcolor[rgb]{0.96,0.65,0.14}{111}\textcolor[rgb]{0.96,0.65,0.14}{^{3}}}$};
\draw (601.94,69.68) node    {$101$};
\draw (569.85,100.56) node    {$010$};
\draw (634.04,99.79) node    {$\textcolor[rgb]{0.29,0.56,0.89}{011}\textcolor[rgb]{0.29,0.56,0.89}{^{2}}$};
\draw (601.94,146.89) node    {$\textcolor[rgb]{0.29,0.56,0.89}{001}\textcolor[rgb]{0.29,0.56,0.89}{^{1}}$};
\draw (271.5,224.68) node    {$\mathbf{\textcolor[rgb]{0.96,0.65,0.14}{100}\textcolor[rgb]{0.96,0.65,0.14}{^{5}}}$};
\draw (311.61,186.07) node    {$\mathbf{\textcolor[rgb]{0.96,0.65,0.14}{110}\textcolor[rgb]{0.96,0.65,0.14}{^{6}}}$};
\draw (271.5,301.89) node    {$\textcolor[rgb]{0.29,0.56,0.89}{000}\textcolor[rgb]{0.29,0.56,0.89}{^{0}}$};
\draw (375.8,193.83) node [anchor=south] [inner sep=0.75pt]    {$\textcolor[rgb]{0.29,0.56,0.89}{111}\textcolor[rgb]{0.29,0.56,0.89}{^{3}}$};
\draw (343.71,224.68) node    {$\mathbf{\textcolor[rgb]{0.96,0.65,0.14}{101}\textcolor[rgb]{0.96,0.65,0.14}{^{4}}}$};
\draw (311.61,256.34) node    {$\mathbf{\textcolor[rgb]{0.96,0.65,0.14}{010}\textcolor[rgb]{0.96,0.65,0.14}{^{7}}}$};
\draw (375.8,254.79) node    {$\textcolor[rgb]{0.29,0.56,0.89}{011}\textcolor[rgb]{0.29,0.56,0.89}{^{2}}$};
\draw (343.71,301.89) node    {$\textcolor[rgb]{0.29,0.56,0.89}{001}\textcolor[rgb]{0.29,0.56,0.89}{^{1}}$};
\draw    (94.96,31.64) -- (126.14,31.66) ;
\draw    (50.92,57.9) -- (65.88,43.63) ;
\draw    (54.84,69.9) -- (94.05,69.9) ;
\draw    (120.62,57.9) -- (132.57,43.68) ;
\draw    (38.34,81.9) -- (38.34,133.94) ;
\draw [color={rgb, 255:red, 0; green, 0; blue, 0 }  ,draw opacity=1 ]   (58.84,146.44) -- (94.05,146.44) ;
\draw    (118.8,134.44) -- (134.39,111.75) ;
\draw [color={rgb, 255:red, 0; green, 0; blue, 0 }  ,draw opacity=1 ]   (49.26,133.94) -- (67.98,112.52) ;
\draw    (94.96,100.32) -- (104.99,100.2)(112.99,100.1) -- (126.14,99.95) ;
\draw    (78.46,43.63) -- (78.46,64.76)(78.46,72.76) -- (78.46,88.52) ;
\draw    (110.55,81.9) -- (110.55,134.44) ;
\draw    (260.47,31.64) -- (291.65,31.66) ;
\draw    (216.43,57.9) -- (231.39,43.63) ;
\draw    (220.35,69.9) -- (259.56,69.9) ;
\draw    (286.13,57.9) -- (298.08,43.68) ;
\draw    (203.85,81.9) -- (203.85,133.94) ;
\draw [color={rgb, 255:red, 245; green, 166; blue, 35 }  ,draw opacity=1 ][line width=0.75]    (224.35,146.22) -- (253.56,145.91) ;
\draw [shift={(255.56,145.89)}, rotate = 539.39] [color={rgb, 255:red, 245; green, 166; blue, 35 }  ,draw opacity=1 ][line width=0.75]    (10.93,-3.29) .. controls (6.95,-1.4) and (3.31,-0.3) .. (0,0) .. controls (3.31,0.3) and (6.95,1.4) .. (10.93,3.29)   ;
\draw    (284.79,133.17) -- (299.76,111.75) ;
\draw [color={rgb, 255:red, 0; green, 0; blue, 0 }  ,draw opacity=1 ]   (214.77,133.94) -- (233.48,112.52) ;
\draw    (260.47,100.32) -- (270.49,100.2)(278.49,100.1) -- (291.65,99.95) ;
\draw    (243.97,43.63) -- (243.97,64.76)(243.97,72.76) -- (243.97,88.52) ;
\draw    (276.06,81.9) -- (276.06,133.17) ;
\draw    (425.97,31.64) -- (457.16,31.66) ;
\draw    (381.94,57.9) -- (396.89,43.63) ;
\draw    (385.86,69.9) -- (425.07,69.9) ;
\draw    (451.64,57.9) -- (463.58,43.68) ;
\draw    (369.36,81.9) -- (369.36,133.94) ;
\draw [color={rgb, 255:red, 74; green, 144; blue, 226 }  ,draw opacity=1 ]   (389.86,146.44) -- (419.07,146.44) ;
\draw [shift={(421.07,146.44)}, rotate = 180] [color={rgb, 255:red, 74; green, 144; blue, 226 }  ,draw opacity=1 ][line width=0.75]    (10.93,-3.29) .. controls (6.95,-1.4) and (3.31,-0.3) .. (0,0) .. controls (3.31,0.3) and (6.95,1.4) .. (10.93,3.29)   ;
\draw [color={rgb, 255:red, 245; green, 166; blue, 35 }  ,draw opacity=1 ][line width=0.75]    (450.16,133.94) -- (463.93,113.9) ;
\draw [shift={(465.06,112.25)}, rotate = 484.5] [color={rgb, 255:red, 245; green, 166; blue, 35 }  ,draw opacity=1 ][line width=0.75]    (10.93,-3.29) .. controls (6.95,-1.4) and (3.31,-0.3) .. (0,0) .. controls (3.31,0.3) and (6.95,1.4) .. (10.93,3.29)   ;
\draw [color={rgb, 255:red, 0; green, 0; blue, 0 }  ,draw opacity=1 ]   (380.28,133.94) -- (398.99,112.52) ;
\draw    (425.97,100.32) -- (434.2,100.22)(442.2,100.13) -- (453.16,100) ;
\draw    (409.47,43.63) -- (409.47,64.76)(409.47,72.76) -- (409.47,88.52) ;
\draw    (441.57,81.9) -- (441.57,133.94) ;
\draw    (586.35,30.98) -- (613.54,30.84) ;
\draw    (542.2,57.68) -- (557.38,43.07) ;
\draw    (546.24,69.68) -- (585.44,69.68) ;
\draw    (611.83,57.68) -- (623.74,43.23) ;
\draw    (529.74,81.68) -- (529.74,134.39) ;
\draw [color={rgb, 255:red, 74; green, 144; blue, 226 }  ,draw opacity=1 ]   (550.24,146.89) -- (579.44,146.89) ;
\draw [shift={(581.44,146.89)}, rotate = 180] [color={rgb, 255:red, 74; green, 144; blue, 226 }  ,draw opacity=1 ][line width=0.75]    (10.93,-3.29) .. controls (6.95,-1.4) and (3.31,-0.3) .. (0,0) .. controls (3.31,0.3) and (6.95,1.4) .. (10.93,3.29)   ;
\draw [color={rgb, 255:red, 74; green, 144; blue, 226 }  ,draw opacity=1 ]   (610.46,134.39) -- (624.39,113.94) ;
\draw [shift={(625.52,112.29)}, rotate = 484.27] [color={rgb, 255:red, 74; green, 144; blue, 226 }  ,draw opacity=1 ][line width=0.75]    (10.93,-3.29) .. controls (6.95,-1.4) and (3.31,-0.3) .. (0,0) .. controls (3.31,0.3) and (6.95,1.4) .. (10.93,3.29)   ;
\draw [color={rgb, 255:red, 0; green, 0; blue, 0 }  ,draw opacity=1 ]   (540.56,134.39) -- (559.46,112.56) ;
\draw    (586.35,100.36) -- (594.58,100.27)(602.58,100.17) -- (613.54,100.04) ;
\draw    (569.85,43.07) -- (569.85,64.55)(569.85,72.55) -- (569.85,88.56) ;
\draw    (601.94,81.68) -- (601.94,134.39) ;
\draw    (332.11,185.96) -- (355.3,185.84) ;
\draw [color={rgb, 255:red, 245; green, 166; blue, 35 }  ,draw opacity=1 ][line width=0.75]    (284.49,212.18) -- (297.19,199.96) ;
\draw [shift={(298.63,198.57)}, rotate = 496.1] [color={rgb, 255:red, 245; green, 166; blue, 35 }  ,draw opacity=1 ][line width=0.75]    (10.93,-3.29) .. controls (6.95,-1.4) and (3.31,-0.3) .. (0,0) .. controls (3.31,0.3) and (6.95,1.4) .. (10.93,3.29)   ;
\draw [color={rgb, 255:red, 245; green, 166; blue, 35 }  ,draw opacity=1 ][line width=0.75]    (294,224.68) -- (323.21,224.68) ;
\draw [shift={(292,224.68)}, rotate = 0] [color={rgb, 255:red, 245; green, 166; blue, 35 }  ,draw opacity=1 ][line width=0.75]    (10.93,-3.29) .. controls (6.95,-1.4) and (3.31,-0.3) .. (0,0) .. controls (3.31,0.3) and (6.95,1.4) .. (10.93,3.29)   ;
\draw [color={rgb, 255:red, 245; green, 166; blue, 35 }  ,draw opacity=1 ][line width=0.75]    (355.28,210.63) -- (365.5,198.23) ;
\draw [shift={(354.01,212.18)}, rotate = 309.49] [color={rgb, 255:red, 245; green, 166; blue, 35 }  ,draw opacity=1 ][line width=0.75]    (10.93,-3.29) .. controls (6.95,-1.4) and (3.31,-0.3) .. (0,0) .. controls (3.31,0.3) and (6.95,1.4) .. (10.93,3.29)   ;
\draw    (271.5,237.18) -- (271.5,289.39) ;
\draw [color={rgb, 255:red, 74; green, 144; blue, 226 }  ,draw opacity=1 ]   (292,301.89) -- (321.21,301.89) ;
\draw [shift={(323.21,301.89)}, rotate = 180] [color={rgb, 255:red, 74; green, 144; blue, 226 }  ,draw opacity=1 ][line width=0.75]    (10.93,-3.29) .. controls (6.95,-1.4) and (3.31,-0.3) .. (0,0) .. controls (3.31,0.3) and (6.95,1.4) .. (10.93,3.29)   ;
\draw [color={rgb, 255:red, 74; green, 144; blue, 226 }  ,draw opacity=1 ]   (352.22,289.39) -- (366.16,268.94) ;
\draw [shift={(367.28,267.29)}, rotate = 484.27] [color={rgb, 255:red, 74; green, 144; blue, 226 }  ,draw opacity=1 ][line width=0.75]    (10.93,-3.29) .. controls (6.95,-1.4) and (3.31,-0.3) .. (0,0) .. controls (3.31,0.3) and (6.95,1.4) .. (10.93,3.29)   ;
\draw [color={rgb, 255:red, 0; green, 0; blue, 0 }  ,draw opacity=1 ]   (282.51,289.39) -- (300.61,268.84) ;
\draw    (332.11,255.84) -- (338.55,255.69)(346.55,255.49) -- (355.3,255.28) ;
\draw [color={rgb, 255:red, 245; green, 166; blue, 35 }  ,draw opacity=1 ][line width=0.75]    (311.61,198.57) -- (311.61,218.8)(311.61,226.8) -- (311.61,241.84) ;
\draw [shift={(311.61,243.84)}, rotate = 270] [color={rgb, 255:red, 245; green, 166; blue, 35 }  ,draw opacity=1 ][line width=0.75]    (10.93,-3.29) .. controls (6.95,-1.4) and (3.31,-0.3) .. (0,0) .. controls (3.31,0.3) and (6.95,1.4) .. (10.93,3.29)   ;
\draw    (343.71,237.18) -- (343.71,289.39) ;

\end{tikzpicture}

%% file: binary_harper.tex
\tikzset{every picture/.style={line width=0.75pt}} 

\begin{tikzpicture}[x=0.75pt,y=0.75pt,yscale=-1,xscale=1]

\draw [line width=0.75]    (136.79,96.8) -- (136.79,47.72) ;
\draw [line width=0.75]    (298.68,96.8) -- (298.68,47.72) ;
\draw [line width=0.75]    (465.54,96.8) -- (465.54,47.72) ;
\draw [line width=0.75]    (625.98,99.35) -- (625.98,50.28) ;
\draw [line width=0.75]    (135.65,267.06) -- (135.65,217.98) ;
\draw [line width=0.75]    (297.58,267.06) -- (297.58,217.98) ;
\draw [line width=0.75]    (463.39,265.78) -- (463.39,216.7) ;
\draw [line width=0.75]    (624.81,268.09) -- (624.81,219.02) ;

\draw (34.59,74.71) node    {$100$};
\draw (74.35,33.82) node    {$110$};
\draw (34.59,156.5) node    {$\mathbf{\textcolor[rgb]{0.96,0.65,0.14}{000}\textcolor[rgb]{0.96,0.65,0.14}{^{0}}}$};
\draw (137.98,41.46) node [anchor=south] [inner sep=0.75pt]    {$111$};
\draw (106.17,74.71) node    {$101$};
\draw (74.35,107.43) node    {$010$};
\draw (137.98,106.61) node    {$011$};
\draw (106.17,156.5) node    {$001$};
\draw (196.48,74.71) node    {$100$};
\draw (236.25,33.82) node    {$110$};
\draw (196.48,156.5) node  [color={rgb, 255:red, 208; green, 2; blue, 27 }  ,opacity=1 ]  {$\textcolor[rgb]{0.29,0.56,0.89}{000}\textcolor[rgb]{0.29,0.56,0.89}{^{0}}$};
\draw (299.88,41.46) node [anchor=south] [inner sep=0.75pt]    {$111$};
\draw (268.06,74.71) node    {$101$};
\draw (236.25,107.43) node    {$010$};
\draw (299.88,106.61) node    {$011$};
\draw (268.06,156.5) node  [color={rgb, 255:red, 208; green, 2; blue, 27 }  ,opacity=1 ]  {$\mathbf{\textcolor[rgb]{0.96,0.65,0.14}{001}\textcolor[rgb]{0.96,0.65,0.14}{^{1}}}$};
\draw (363.34,74.71) node    {$100$};
\draw (403.11,33.82) node    {$110$};
\draw (363.34,156.5) node  [color={rgb, 255:red, 208; green, 2; blue, 27 }  ,opacity=1 ]  {$\textcolor[rgb]{0.29,0.56,0.89}{000}\textcolor[rgb]{0.29,0.56,0.89}{^{0}}$};
\draw (466.74,41.46) node [anchor=south] [inner sep=0.75pt]    {$111$};
\draw (434.92,74.71) node    {$101$};
\draw (403.11,107.43) node  [color={rgb, 255:red, 208; green, 2; blue, 27 }  ,opacity=1 ]  {$\mathbf{\textcolor[rgb]{0.96,0.65,0.14}{010}\textcolor[rgb]{0.96,0.65,0.14}{^{2}}}$};
\draw (466.74,106.61) node    {$011$};
\draw (434.92,156.5) node  [color={rgb, 255:red, 208; green, 2; blue, 27 }  ,opacity=1 ]  {$\textcolor[rgb]{0.29,0.56,0.89}{001}\textcolor[rgb]{0.29,0.56,0.89}{^{1}}$};
\draw (523.78,77.27) node    {$100$};
\draw (563.55,36.38) node    {$110$};
\draw (523.78,158.24) node    {$\textcolor[rgb]{0.29,0.56,0.89}{000}\textcolor[rgb]{0.29,0.56,0.89}{^{0}}$};
\draw (627.18,44.02) node [anchor=south] [inner sep=0.75pt]    {$111$};
\draw (595.36,77.27) node    {$101$};
\draw (563.55,109.17) node  [color={rgb, 255:red, 208; green, 2; blue, 27 }  ,opacity=1 ]  {$\textcolor[rgb]{0.29,0.56,0.89}{010}\textcolor[rgb]{0.29,0.56,0.89}{^{2}}$};
\draw (627.18,108.35) node  [color={rgb, 255:red, 208; green, 2; blue, 27 }  ,opacity=1 ]  {$\mathbf{\textcolor[rgb]{0.96,0.65,0.14}{011}\textcolor[rgb]{0.96,0.65,0.14}{^{3}}}$};
\draw (595.36,159.06) node  [color={rgb, 255:red, 208; green, 2; blue, 27 }  ,opacity=1 ]  {$\textcolor[rgb]{0.29,0.56,0.89}{001}\textcolor[rgb]{0.29,0.56,0.89}{^{1}}$};
\draw (33.45,244.97) node  [color={rgb, 255:red, 208; green, 2; blue, 27 }  ,opacity=1 ]  {$\mathbf{\textcolor[rgb]{0.96,0.65,0.14}{100}\textcolor[rgb]{0.96,0.65,0.14}{^{4}}}$};
\draw (73.22,204.08) node    {$110$};
\draw (33.45,326.76) node  [color={rgb, 255:red, 208; green, 2; blue, 27 }  ,opacity=1 ]  {$\textcolor[rgb]{0.29,0.56,0.89}{000}\textcolor[rgb]{0.29,0.56,0.89}{^{0}}$};
\draw (136.85,211.72) node [anchor=south] [inner sep=0.75pt]    {$111$};
\draw (105.03,244.97) node    {$101$};
\draw (73.22,276.87) node  [color={rgb, 255:red, 208; green, 2; blue, 27 }  ,opacity=1 ]  {$\textcolor[rgb]{0.29,0.56,0.89}{010}\textcolor[rgb]{0.29,0.56,0.89}{^{2}}$};
\draw (136.85,276.87) node  [color={rgb, 255:red, 208; green, 2; blue, 27 }  ,opacity=1 ]  {$\textcolor[rgb]{0.29,0.56,0.89}{011}\textcolor[rgb]{0.29,0.56,0.89}{^{3}}$};
\draw (105.03,326.76) node  [color={rgb, 255:red, 208; green, 2; blue, 27 }  ,opacity=1 ]  {$\textcolor[rgb]{0.29,0.56,0.89}{001}\textcolor[rgb]{0.29,0.56,0.89}{^{1}}$};
\draw (195.38,244.97) node    {$\textcolor[rgb]{0.29,0.56,0.89}{100}\textcolor[rgb]{0.29,0.56,0.89}{^{4}}$};
\draw (235.14,204.08) node    {$110$};
\draw (195.38,326.76) node  [color={rgb, 255:red, 208; green, 2; blue, 27 }  ,opacity=1 ]  {$\textcolor[rgb]{0.29,0.56,0.89}{000}\textcolor[rgb]{0.29,0.56,0.89}{^{0}}$};
\draw (298.77,211.72) node [anchor=south] [inner sep=0.75pt]    {$111$};
\draw (266.96,244.97) node    {$\mathbf{\textcolor[rgb]{0.96,0.65,0.14}{101}\textcolor[rgb]{0.96,0.65,0.14}{^{5}}}$};
\draw (235.14,277.69) node  [color={rgb, 255:red, 208; green, 2; blue, 27 }  ,opacity=1 ]  {$\textcolor[rgb]{0.29,0.56,0.89}{010}\textcolor[rgb]{0.29,0.56,0.89}{^{2}}$};
\draw (298.77,277.69) node    {$\textcolor[rgb]{0.29,0.56,0.89}{011}\textcolor[rgb]{0.29,0.56,0.89}{^{3}}$};
\draw (266.96,326.76) node  [color={rgb, 255:red, 208; green, 2; blue, 27 }  ,opacity=1 ]  {$\textcolor[rgb]{0.29,0.56,0.89}{001}\textcolor[rgb]{0.29,0.56,0.89}{^{1}}$};
\draw (361.19,243.69) node    {$\textcolor[rgb]{0.29,0.56,0.89}{100}\textcolor[rgb]{0.29,0.56,0.89}{^{4}}$};
\draw (400.96,202.8) node    {$\mathbf{\textcolor[rgb]{0.96,0.65,0.14}{110}\textcolor[rgb]{0.96,0.65,0.14}{^{6}}}$};
\draw (361.19,325.48) node    {$\textcolor[rgb]{0.29,0.56,0.89}{000}\textcolor[rgb]{0.29,0.56,0.89}{^{0}}$};
\draw (464.59,210.45) node [anchor=south] [inner sep=0.75pt]    {$111$};
\draw (432.77,243.69) node    {$\textcolor[rgb]{0.29,0.56,0.89}{101}\textcolor[rgb]{0.29,0.56,0.89}{^{5}}$};
\draw (400.96,275.59) node  [color={rgb, 255:red, 208; green, 2; blue, 27 }  ,opacity=1 ]  {$\textcolor[rgb]{0.29,0.56,0.89}{010}\textcolor[rgb]{0.29,0.56,0.89}{^{2}}$};
\draw (464.59,275.59) node  [color={rgb, 255:red, 208; green, 2; blue, 27 }  ,opacity=1 ]  {$\textcolor[rgb]{0.29,0.56,0.89}{011}\textcolor[rgb]{0.29,0.56,0.89}{^{3}}$};
\draw (432.77,325.48) node  [color={rgb, 255:red, 208; green, 2; blue, 27 }  ,opacity=1 ]  {$\textcolor[rgb]{0.29,0.56,0.89}{001}\textcolor[rgb]{0.29,0.56,0.89}{^{1}}$};
\draw (522.6,246.01) node    {$\textcolor[rgb]{0.29,0.56,0.89}{100}\textcolor[rgb]{0.29,0.56,0.89}{^{4}}$};
\draw (562.37,205.11) node    {$\textcolor[rgb]{0.29,0.56,0.89}{110}\textcolor[rgb]{0.29,0.56,0.89}{^{6}}$};
\draw (522.6,327.8) node    {$\textcolor[rgb]{0.29,0.56,0.89}{000}\textcolor[rgb]{0.29,0.56,0.89}{^{0}}$};
\draw (626,212.85) node [anchor=south] [inner sep=0.75pt]    {$\mathbf{\textcolor[rgb]{0.96,0.65,0.14}{111}\textcolor[rgb]{0.96,0.65,0.14}{^{7}}}$};
\draw (594.19,246.01) node    {$\textcolor[rgb]{0.29,0.56,0.89}{101}\textcolor[rgb]{0.29,0.56,0.89}{^{5}}$};
\draw (562.37,277.9) node  [color={rgb, 255:red, 208; green, 2; blue, 27 }  ,opacity=1 ]  {$\textcolor[rgb]{0.29,0.56,0.89}{010}\textcolor[rgb]{0.29,0.56,0.89}{^{2}}$};
\draw (626,277.9) node  [color={rgb, 255:red, 208; green, 2; blue, 27 }  ,opacity=1 ]  {$\textcolor[rgb]{0.29,0.56,0.89}{011}\textcolor[rgb]{0.29,0.56,0.89}{^{3}}$};
\draw (594.19,327.8) node  [color={rgb, 255:red, 208; green, 2; blue, 27 }  ,opacity=1 ]  {$\textcolor[rgb]{0.29,0.56,0.89}{001}\textcolor[rgb]{0.29,0.56,0.89}{^{1}}$};
\draw    (90.85,33.83) -- (121.48,33.85) ;
\draw    (46.26,62.71) -- (62.68,45.82) ;
\draw    (51.09,74.71) -- (89.67,74.71) ;
\draw    (115.51,62.71) -- (128.63,45.86) ;
\draw    (34.59,86.71) -- (34.59,142) ;
\draw [color={rgb, 255:red, 0; green, 0; blue, 0 }  ,draw opacity=1 ]   (55.09,156.5) -- (89.67,156.5) ;
\draw    (113.82,144.5) -- (130.33,118.61) ;
\draw [color={rgb, 255:red, 0; green, 0; blue, 0 }  ,draw opacity=1 ]   (46.34,142) -- (64.63,119.43) ;
\draw    (90.85,107.22) -- (100.63,107.09)(108.63,106.99) -- (121.48,106.82) ;
\draw    (74.35,45.82) -- (74.35,69.6)(74.35,77.6) -- (74.35,95.43) ;
\draw    (106.17,86.71) -- (106.17,144.5) ;
\draw    (252.75,33.83) -- (283.38,33.85) ;
\draw    (208.15,62.71) -- (224.58,45.82) ;
\draw    (212.98,74.71) -- (251.56,74.71) ;
\draw    (277.41,62.71) -- (290.53,45.86) ;
\draw    (196.48,86.71) -- (196.48,144) ;
\draw [color={rgb, 255:red, 0; green, 0; blue, 0 }  ,draw opacity=1 ]   (216.98,156.5) -- (247.56,156.5) ;
\draw    (276.03,144) -- (292.22,118.61) ;
\draw [color={rgb, 255:red, 0; green, 0; blue, 0 }  ,draw opacity=1 ]   (206.61,144) -- (226.53,119.43) ;
\draw    (252.75,107.22) -- (262.53,107.09)(270.53,106.99) -- (283.38,106.82) ;
\draw    (236.25,45.82) -- (236.25,69.6)(236.25,77.6) -- (236.25,95.43) ;
\draw    (268.06,86.71) -- (268.06,144) ;
\draw    (419.61,33.83) -- (450.24,33.85) ;
\draw    (375.01,62.71) -- (391.44,45.82) ;
\draw    (379.84,74.71) -- (418.42,74.71) ;
\draw    (444.27,62.71) -- (457.39,45.86) ;
\draw    (363.34,86.71) -- (363.34,144) ;
\draw [color={rgb, 255:red, 0; green, 0; blue, 0 }  ,draw opacity=1 ]   (383.84,156.5) -- (414.42,156.5) ;
\draw    (442.89,144) -- (459.08,118.61) ;
\draw [color={rgb, 255:red, 0; green, 0; blue, 0 }  ,draw opacity=1 ]   (373.47,144) -- (392.98,119.93) ;
\draw    (423.61,107.17) -- (431.59,107.06)(439.59,106.96) -- (450.24,106.82) ;
\draw    (403.11,45.82) -- (403.11,69.32)(403.11,77.32) -- (403.11,94.93) ;
\draw    (434.92,86.71) -- (434.92,144) ;
\draw    (580.05,36.39) -- (610.68,36.41) ;
\draw    (535.45,65.27) -- (551.88,48.38) ;
\draw    (540.28,77.27) -- (578.86,77.27) ;
\draw    (604.71,65.27) -- (617.83,48.42) ;
\draw    (523.78,89.27) -- (523.78,143.74) ;
\draw [color={rgb, 255:red, 0; green, 0; blue, 0 }  ,draw opacity=1 ]   (544.28,158.47) -- (574.86,158.82) ;
\draw    (603.21,146.56) -- (619.33,120.85) ;
\draw [color={rgb, 255:red, 0; green, 0; blue, 0 }  ,draw opacity=1 ]   (535.53,143.74) -- (553.42,121.67) ;
\draw    (584.05,108.9) -- (590.23,108.82)(598.23,108.72) -- (606.68,108.61) ;
\draw    (563.55,48.38) -- (563.55,71.42)(563.55,79.42) -- (563.55,96.67) ;
\draw    (595.36,89.27) -- (595.36,146.56) ;
\draw    (89.72,204.09) -- (120.35,204.11) ;
\draw    (45.61,232.47) -- (61.55,216.08) ;
\draw    (53.95,244.97) -- (88.53,244.97) ;
\draw    (114.38,232.97) -- (127.5,216.12) ;
\draw    (33.45,257.47) -- (33.45,314.26) ;
\draw [color={rgb, 255:red, 0; green, 0; blue, 0 }  ,draw opacity=1 ]   (53.95,326.76) -- (84.53,326.76) ;
\draw    (113,314.26) -- (128.87,289.37) ;
\draw [color={rgb, 255:red, 0; green, 0; blue, 0 }  ,draw opacity=1 ]   (43.41,314.26) -- (63.25,289.37) ;
\draw    (93.72,276.87) -- (99.9,276.87)(107.9,276.87) -- (116.35,276.87) ;
\draw    (73.22,216.08) -- (73.22,239.12)(73.22,247.12) -- (73.22,264.37) ;
\draw    (105.03,256.97) -- (105.03,314.26) ;
\draw    (251.64,204.09) -- (282.27,204.11) ;
\draw    (207.53,232.47) -- (223.47,216.08) ;
\draw    (215.88,244.97) -- (246.46,244.97) ;
\draw    (276.69,232.47) -- (289.43,216.12) ;
\draw    (195.38,257.47) -- (195.38,314.26) ;
\draw [color={rgb, 255:red, 0; green, 0; blue, 0 }  ,draw opacity=1 ]   (215.88,326.76) -- (246.46,326.76) ;
\draw    (275.06,314.26) -- (290.67,290.19) ;
\draw [color={rgb, 255:red, 0; green, 0; blue, 0 }  ,draw opacity=1 ]   (205.51,314.26) -- (225.01,290.19) ;
\draw    (255.64,277.69) -- (261.82,277.69)(269.82,277.69) -- (278.27,277.69) ;
\draw    (235.14,216.08) -- (235.14,239.58)(235.14,247.58) -- (235.14,265.19) ;
\draw    (266.96,257.47) -- (266.96,314.26) ;
\draw    (421.46,202.82) -- (448.09,202.83) ;
\draw    (373.35,231.19) -- (388.8,215.3) ;
\draw    (381.69,243.69) -- (412.27,243.69) ;
\draw    (442.51,231.19) -- (455.24,214.85) ;
\draw    (361.19,256.19) -- (361.19,310.98) ;
\draw [color={rgb, 255:red, 0; green, 0; blue, 0 }  ,draw opacity=1 ]   (381.69,325.48) -- (412.27,325.48) ;
\draw    (440.74,312.98) -- (456.62,288.09) ;
\draw [color={rgb, 255:red, 0; green, 0; blue, 0 }  ,draw opacity=1 ]   (372.75,310.98) -- (391,288.09) ;
\draw    (421.46,275.59) -- (427.64,275.59)(435.64,275.59) -- (444.09,275.59) ;
\draw    (400.96,215.3) -- (400.96,238.06)(400.96,246.06) -- (400.96,263.09) ;
\draw    (432.77,256.19) -- (432.77,312.98) ;
\draw    (582.87,205) -- (605.5,204.87) ;
\draw    (534.76,233.51) -- (550.22,217.61) ;
\draw    (543.1,246.01) -- (573.69,246.01) ;
\draw    (603.82,233.51) -- (616.36,217.25) ;
\draw    (522.6,258.51) -- (522.6,313.3) ;
\draw [color={rgb, 255:red, 0; green, 0; blue, 0 }  ,draw opacity=1 ]   (543.1,327.8) -- (573.69,327.8) ;
\draw    (602.16,315.3) -- (618.03,290.4) ;
\draw [color={rgb, 255:red, 0; green, 0; blue, 0 }  ,draw opacity=1 ]   (534.16,313.3) -- (552.41,290.4) ;
\draw    (582.87,277.9) -- (589.05,277.9)(597.05,277.9) -- (605.5,277.9) ;
\draw    (562.37,217.61) -- (562.37,240.37)(562.37,248.37) -- (562.37,265.4) ;
\draw    (594.19,258.51) -- (594.19,315.3) ;

\end{tikzpicture}

%% file: brg_harper.tex
\tikzset{every picture/.style={line width=0.75pt}} 

\begin{tikzpicture}[x=0.75pt,y=0.75pt,yscale=-1,xscale=1]

\draw [line width=0.75]    (148.34,94.66) -- (148.34,46.62) ;
\draw [line width=0.75]    (311.93,94.66) -- (311.93,46.62) ;
\draw [line width=0.75]    (470.52,94.66) -- (470.52,46.62) ;
\draw [line width=0.75]    (628.54,93.59) -- (628.54,45.55) ;
\draw [line width=0.75]    (143.14,270.19) -- (143.14,222.15) ;
\draw [line width=0.75]    (306.73,270.19) -- (306.73,222.15) ;
\draw [line width=0.75]    (470.96,266.94) -- (470.96,218.9) ;
\draw [line width=0.75]    (626.93,272.54) -- (626.93,224.51) ;

\draw (45.66,73.04) node    {$100$};
\draw (85.62,33.01) node    {$110$};
\draw (45.66,153.1) node    {$\mathbf{\textcolor[rgb]{0.96,0.65,0.14}{000}\textcolor[rgb]{0.96,0.65,0.14}{^{0}}}$};
\draw (149.54,40.65) node [anchor=south] [inner sep=0.75pt]    {$111$};
\draw (117.58,73.04) node    {$101$};
\draw (85.62,105.06) node    {$010$};
\draw (149.54,104.26) node    {$011$};
\draw (117.58,153.1) node    {$001$};
\draw (209.26,73.04) node    {$100$};
\draw (249.21,33.01) node    {$110$};
\draw (209.26,153.1) node  [color={rgb, 255:red, 208; green, 2; blue, 27 }  ,opacity=1 ]  {$\textcolor[rgb]{0.29,0.56,0.89}{000}\textcolor[rgb]{0.29,0.56,0.89}{^{0}}$};
\draw (313.13,40.65) node [anchor=south] [inner sep=0.75pt]    {$111$};
\draw (281.17,73.04) node    {$101$};
\draw (249.21,105.06) node    {$010$};
\draw (313.13,104.26) node    {$011$};
\draw (281.17,153.1) node  [color={rgb, 255:red, 208; green, 2; blue, 27 }  ,opacity=1 ]  {$\mathbf{\textcolor[rgb]{0.96,0.65,0.14}{001}\textcolor[rgb]{0.96,0.65,0.14}{^{1}}}$};
\draw (367.85,73.04) node    {$100$};
\draw (407.8,33.01) node    {$110$};
\draw (367.85,153.1) node  [color={rgb, 255:red, 208; green, 2; blue, 27 }  ,opacity=1 ]  {$\textcolor[rgb]{0.29,0.56,0.89}{000}\textcolor[rgb]{0.29,0.56,0.89}{^{0}}$};
\draw (471.72,40.65) node [anchor=south] [inner sep=0.75pt]    {$111$};
\draw (439.76,73.04) node    {$101$};
\draw (407.8,105.06) node  [color={rgb, 255:red, 208; green, 2; blue, 27 }  ,opacity=1 ]  {$\textcolor[rgb]{0,0,0}{010}\textcolor[rgb]{0,0,0}{}$};
\draw (471.72,104.26) node    {$\mathbf{\textcolor[rgb]{0.96,0.65,0.14}{011}\textcolor[rgb]{0.96,0.65,0.14}{^{2}}}$};
\draw (439.76,153.1) node  [color={rgb, 255:red, 208; green, 2; blue, 27 }  ,opacity=1 ]  {$\textcolor[rgb]{0.29,0.56,0.89}{001}\textcolor[rgb]{0.29,0.56,0.89}{^{1}}$};
\draw (525.87,71.97) node    {$100$};
\draw (565.82,31.94) node    {$110$};
\draw (525.87,152.04) node    {$\textcolor[rgb]{0.29,0.56,0.89}{000}\textcolor[rgb]{0.29,0.56,0.89}{^{0}}$};
\draw (629.74,39.59) node [anchor=south] [inner sep=0.75pt]    {$111$};
\draw (597.78,71.97) node    {$101$};
\draw (565.82,103.2) node  [color={rgb, 255:red, 208; green, 2; blue, 27 }  ,opacity=1 ]  {$\mathbf{\textcolor[rgb]{0.96,0.65,0.14}{010}\textcolor[rgb]{0.96,0.65,0.14}{^{3}}}$};
\draw (629.74,103.2) node  [color={rgb, 255:red, 208; green, 2; blue, 27 }  ,opacity=1 ]  {$\textcolor[rgb]{0.29,0.56,0.89}{011}\textcolor[rgb]{0.29,0.56,0.89}{^{2}}$};
\draw (597.78,152.04) node  [color={rgb, 255:red, 208; green, 2; blue, 27 }  ,opacity=1 ]  {$\textcolor[rgb]{0.29,0.56,0.89}{001}\textcolor[rgb]{0.29,0.56,0.89}{^{1}}$};
\draw (40.46,248.58) node  [color={rgb, 255:red, 208; green, 2; blue, 27 }  ,opacity=1 ]  {$\textcolor[rgb]{0,0,0}{100}$};
\draw (80.41,208.54) node    {$\mathbf{\textcolor[rgb]{0.96,0.65,0.14}{110}\textcolor[rgb]{0.96,0.65,0.14}{^{4}}}$};
\draw (40.46,328.64) node  [color={rgb, 255:red, 208; green, 2; blue, 27 }  ,opacity=1 ]  {$\textcolor[rgb]{0.29,0.56,0.89}{000}\textcolor[rgb]{0.29,0.56,0.89}{^{0}}$};
\draw (144.34,216.19) node [anchor=south] [inner sep=0.75pt]    {$111$};
\draw (112.38,248.58) node    {$101$};
\draw (80.41,280.6) node  [color={rgb, 255:red, 208; green, 2; blue, 27 }  ,opacity=1 ]  {$\textcolor[rgb]{0.29,0.56,0.89}{010}\textcolor[rgb]{0.29,0.56,0.89}{^{3}}$};
\draw (144.34,279.8) node  [color={rgb, 255:red, 208; green, 2; blue, 27 }  ,opacity=1 ]  {$\textcolor[rgb]{0.29,0.56,0.89}{011}\textcolor[rgb]{0.29,0.56,0.89}{^{2}}$};
\draw (112.38,328.64) node  [color={rgb, 255:red, 208; green, 2; blue, 27 }  ,opacity=1 ]  {$\textcolor[rgb]{0.29,0.56,0.89}{001}\textcolor[rgb]{0.29,0.56,0.89}{^{1}}$};
\draw (204.05,248.58) node    {$\textcolor[rgb]{0,0,0}{100}$};
\draw (244,208.54) node    {$\textcolor[rgb]{0.29,0.56,0.89}{110}\textcolor[rgb]{0.29,0.56,0.89}{^{4}}$};
\draw (204.05,328.64) node  [color={rgb, 255:red, 208; green, 2; blue, 27 }  ,opacity=1 ]  {$\textcolor[rgb]{0.29,0.56,0.89}{000}\textcolor[rgb]{0.29,0.56,0.89}{^{0}}$};
\draw (307.92,216.29) node [anchor=south] [inner sep=0.75pt]    {$\mathbf{\textcolor[rgb]{0.96,0.65,0.14}{111}\textcolor[rgb]{0.96,0.65,0.14}{^{5}}}$};
\draw (275.96,248.58) node    {$\textcolor[rgb]{0,0,0}{101}$};
\draw (244,280.6) node  [color={rgb, 255:red, 208; green, 2; blue, 27 }  ,opacity=1 ]  {$\textcolor[rgb]{0.29,0.56,0.89}{010}\textcolor[rgb]{0.29,0.56,0.89}{^{3}}$};
\draw (307.92,279.8) node    {$\textcolor[rgb]{0.29,0.56,0.89}{011}\textcolor[rgb]{0.29,0.56,0.89}{^{2}}$};
\draw (275.96,328.64) node  [color={rgb, 255:red, 208; green, 2; blue, 27 }  ,opacity=1 ]  {$\textcolor[rgb]{0.29,0.56,0.89}{001}\textcolor[rgb]{0.29,0.56,0.89}{^{1}}$};
\draw (368.28,245.32) node    {$\textcolor[rgb]{0,0,0}{100}$};
\draw (408.23,205.29) node    {$\textcolor[rgb]{0.29,0.56,0.89}{110}\textcolor[rgb]{0.29,0.56,0.89}{^{4}}$};
\draw (368.28,325.39) node    {$\textcolor[rgb]{0.29,0.56,0.89}{000}\textcolor[rgb]{0.29,0.56,0.89}{^{0}}$};
\draw (472.15,213.03) node [anchor=south] [inner sep=0.75pt]    {$\textcolor[rgb]{0.29,0.56,0.89}{111}\textcolor[rgb]{0.29,0.56,0.89}{^{5}}$};
\draw (440.19,245.32) node    {$\mathbf{\textcolor[rgb]{0.96,0.65,0.14}{101}\textcolor[rgb]{0.96,0.65,0.14}{^{6}}}$};
\draw (408.23,276.55) node  [color={rgb, 255:red, 208; green, 2; blue, 27 }  ,opacity=1 ]  {$\textcolor[rgb]{0.29,0.56,0.89}{010}\textcolor[rgb]{0.29,0.56,0.89}{^{3}}$};
\draw (472.15,276.55) node  [color={rgb, 255:red, 208; green, 2; blue, 27 }  ,opacity=1 ]  {$\textcolor[rgb]{0.29,0.56,0.89}{011}\textcolor[rgb]{0.29,0.56,0.89}{^{2}}$};
\draw (440.19,325.39) node  [color={rgb, 255:red, 208; green, 2; blue, 27 }  ,opacity=1 ]  {$\textcolor[rgb]{0.29,0.56,0.89}{001}\textcolor[rgb]{0.29,0.56,0.89}{^{1}}$};
\draw (524.26,250.93) node    {$\mathbf{\textcolor[rgb]{0.96,0.65,0.14}{100}\textcolor[rgb]{0.96,0.65,0.14}{^{7}}}$};
\draw (564.21,210.9) node    {$\textcolor[rgb]{0.29,0.56,0.89}{110}\textcolor[rgb]{0.29,0.56,0.89}{^{4}}$};
\draw (524.26,330.99) node    {$\textcolor[rgb]{0.29,0.56,0.89}{000}\textcolor[rgb]{0.29,0.56,0.89}{^{0}}$};
\draw (628.13,218.64) node [anchor=south] [inner sep=0.75pt]    {$\textcolor[rgb]{0.29,0.56,0.89}{111}\textcolor[rgb]{0.29,0.56,0.89}{^{5}}$};
\draw (596.17,250.93) node    {$\textcolor[rgb]{0.29,0.56,0.89}{101}\textcolor[rgb]{0.29,0.56,0.89}{^{6}}$};
\draw (564.21,282.15) node  [color={rgb, 255:red, 208; green, 2; blue, 27 }  ,opacity=1 ]  {$\textcolor[rgb]{0.29,0.56,0.89}{010}\textcolor[rgb]{0.29,0.56,0.89}{^{3}}$};
\draw (628.13,282.15) node  [color={rgb, 255:red, 208; green, 2; blue, 27 }  ,opacity=1 ]  {$\textcolor[rgb]{0.29,0.56,0.89}{011}\textcolor[rgb]{0.29,0.56,0.89}{^{2}}$};
\draw (596.17,330.99) node  [color={rgb, 255:red, 208; green, 2; blue, 27 }  ,opacity=1 ]  {$\textcolor[rgb]{0.29,0.56,0.89}{001}\textcolor[rgb]{0.29,0.56,0.89}{^{1}}$};
\draw    (102.12,33.02) -- (133.04,33.04) ;
\draw    (57.64,61.04) -- (73.64,45.01) ;
\draw    (62.16,73.04) -- (101.08,73.04) ;
\draw    (127.17,61.04) -- (139.95,45.05) ;
\draw    (45.66,85.04) -- (45.66,138.6) ;
\draw [color={rgb, 255:red, 0; green, 0; blue, 0 }  ,draw opacity=1 ]   (66.16,153.1) -- (101.08,153.1) ;
\draw    (125.43,141.1) -- (141.68,116.26) ;
\draw [color={rgb, 255:red, 0; green, 0; blue, 0 }  ,draw opacity=1 ]   (57.72,138.6) -- (75.64,117.06) ;
\draw    (102.12,104.86) -- (112.03,104.73)(120.03,104.63) -- (133.04,104.47) ;
\draw    (85.62,45.01) -- (85.62,67.92)(85.62,75.92) -- (85.62,93.06) ;
\draw    (117.58,85.04) -- (117.58,141.1) ;
\draw    (265.71,33.02) -- (296.63,33.04) ;
\draw    (221.23,61.04) -- (237.23,45.01) ;
\draw    (225.76,73.04) -- (264.67,73.04) ;
\draw    (290.76,61.04) -- (303.54,45.05) ;
\draw    (209.26,85.04) -- (209.26,140.6) ;
\draw [color={rgb, 255:red, 0; green, 0; blue, 0 }  ,draw opacity=1 ]   (229.76,153.1) -- (260.67,153.1) ;
\draw    (289.35,140.6) -- (305.28,116.26) ;
\draw [color={rgb, 255:red, 0; green, 0; blue, 0 }  ,draw opacity=1 ]   (219.65,140.6) -- (239.23,117.06) ;
\draw    (265.71,104.86) -- (275.63,104.73)(283.63,104.63) -- (296.63,104.47) ;
\draw    (249.21,45.01) -- (249.21,67.92)(249.21,75.92) -- (249.21,93.06) ;
\draw    (281.17,85.04) -- (281.17,140.6) ;
\draw    (424.3,33.02) -- (455.22,33.04) ;
\draw    (379.82,61.04) -- (395.82,45.01) ;
\draw    (384.35,73.04) -- (423.26,73.04) ;
\draw    (449.35,61.04) -- (462.13,45.05) ;
\draw    (367.85,85.04) -- (367.85,140.6) ;
\draw [color={rgb, 255:red, 0; green, 0; blue, 0 }  ,draw opacity=1 ]   (388.35,153.1) -- (419.26,153.1) ;
\draw    (447.94,140.6) -- (463.54,116.76) ;
\draw [color={rgb, 255:red, 0; green, 0; blue, 0 }  ,draw opacity=1 ]   (378.24,140.6) -- (396.99,118.06) ;
\draw    (428.3,104.81) -- (434.62,104.73)(442.62,104.63) -- (451.22,104.52) ;
\draw    (407.8,45.01) -- (407.8,67.36)(407.8,75.36) -- (407.8,92.06) ;
\draw    (439.76,85.04) -- (439.76,140.6) ;
\draw    (582.32,31.95) -- (613.24,31.98) ;
\draw    (537.85,59.97) -- (553.85,43.94) ;
\draw    (542.37,71.97) -- (581.28,71.97) ;
\draw    (607.37,59.97) -- (620.15,43.99) ;
\draw    (525.87,83.97) -- (525.87,137.54) ;
\draw [color={rgb, 255:red, 0; green, 0; blue, 0 }  ,draw opacity=1 ]   (546.37,152.04) -- (577.28,152.04) ;
\draw    (605.96,139.54) -- (621.56,115.7) ;
\draw [color={rgb, 255:red, 0; green, 0; blue, 0 }  ,draw opacity=1 ]   (537.73,137.54) -- (555.6,115.7) ;
\draw    (586.32,103.2) -- (592.63,103.2)(600.63,103.2) -- (609.24,103.2) ;
\draw    (565.82,43.94) -- (565.82,66.12)(565.82,74.12) -- (565.82,90.7) ;
\draw    (597.78,83.97) -- (597.78,139.54) ;
\draw    (100.91,208.56) -- (127.84,208.58) ;
\draw    (52.44,236.58) -- (67.94,221.04) ;
\draw    (56.96,248.58) -- (95.88,248.58) ;
\draw    (121.97,236.58) -- (134.74,220.59) ;
\draw    (40.46,260.58) -- (40.46,316.14) ;
\draw [color={rgb, 255:red, 0; green, 0; blue, 0 }  ,draw opacity=1 ]   (60.96,328.64) -- (91.88,328.64) ;
\draw    (120.56,316.14) -- (136.16,292.3) ;
\draw [color={rgb, 255:red, 0; green, 0; blue, 0 }  ,draw opacity=1 ]   (50.86,316.14) -- (70.02,293.1) ;
\draw    (100.91,280.35) -- (107.23,280.27)(115.23,280.17) -- (123.84,280.06) ;
\draw    (80.41,221.04) -- (80.41,243.39)(80.41,251.39) -- (80.41,268.1) ;
\draw    (112.38,260.58) -- (112.38,316.14) ;
\draw    (264.5,208.43) -- (287.42,208.3) ;
\draw    (216.03,236.58) -- (231.53,221.04) ;
\draw    (220.55,248.58) -- (259.46,248.58) ;
\draw    (285.46,236.58) -- (298.03,220.69) ;
\draw    (204.05,260.58) -- (204.05,316.14) ;
\draw [color={rgb, 255:red, 0; green, 0; blue, 0 }  ,draw opacity=1 ]   (224.55,328.64) -- (255.46,328.64) ;
\draw    (284.14,316.14) -- (299.74,292.3) ;
\draw [color={rgb, 255:red, 0; green, 0; blue, 0 }  ,draw opacity=1 ]   (214.45,316.14) -- (233.61,293.1) ;
\draw    (264.5,280.35) -- (270.82,280.27)(278.82,280.17) -- (287.42,280.06) ;
\draw    (244,221.04) -- (244,243.39)(244,251.39) -- (244,268.1) ;
\draw    (275.96,260.58) -- (275.96,316.14) ;
\draw    (428.73,205.18) -- (451.65,205.05) ;
\draw    (380.26,233.32) -- (395.76,217.79) ;
\draw    (384.78,245.32) -- (419.69,245.32) ;
\draw    (450.09,232.82) -- (462.26,217.43) ;
\draw    (368.28,257.32) -- (368.28,310.89) ;
\draw [color={rgb, 255:red, 0; green, 0; blue, 0 }  ,draw opacity=1 ]   (388.78,325.39) -- (419.69,325.39) ;
\draw    (448.37,312.89) -- (463.97,289.05) ;
\draw [color={rgb, 255:red, 0; green, 0; blue, 0 }  ,draw opacity=1 ]   (380.14,310.89) -- (398.01,289.05) ;
\draw    (428.73,276.55) -- (435.04,276.55)(443.04,276.55) -- (451.65,276.55) ;
\draw    (408.23,217.79) -- (408.23,239.69)(408.23,247.69) -- (408.23,264.05) ;
\draw    (440.19,257.82) -- (440.19,312.89) ;
\draw    (584.71,210.78) -- (607.63,210.65) ;
\draw    (536.73,238.43) -- (551.73,223.4) ;
\draw    (544.76,250.93) -- (575.67,250.93) ;
\draw    (606.06,238.43) -- (618.24,223.04) ;
\draw    (524.26,263.43) -- (524.26,316.49) ;
\draw [color={rgb, 255:red, 0; green, 0; blue, 0 }  ,draw opacity=1 ]   (544.76,330.99) -- (575.67,330.99) ;
\draw    (604.35,318.49) -- (619.95,294.65) ;
\draw [color={rgb, 255:red, 0; green, 0; blue, 0 }  ,draw opacity=1 ]   (536.12,316.49) -- (553.98,294.65) ;
\draw    (584.71,282.15) -- (591.02,282.15)(599.02,282.15) -- (607.63,282.15) ;
\draw    (564.21,223.4) -- (564.21,245.3)(564.21,253.3) -- (564.21,269.65) ;
\draw    (596.17,263.43) -- (596.17,318.49) ;

\end{tikzpicture}